%% file: Main.tex
\ifdefined\pdfobjcompresslevel
\fi
\documentclass[dvipsnames]{article}

\usepackage[final]{openbmb_2025}

\usepackage{amsmath}
\usepackage{amssymb}
\usepackage{mathtools}
\usepackage{amsthm}
\tcbuselibrary{breakable}
\usepackage{multirow}
\usepackage{makecell}
\usepackage{subcaption}
\usepackage{wrapfig}
\usepackage{graphicx}
\usepackage{pifont}
\usepackage{tabularx}
\usepackage{longtable}
\usepackage[utf8]{inputenc} 
\usepackage[T1]{fontenc}    
\usepackage{url}            
\usepackage{fvextra}        
\usepackage{booktabs}       
\usepackage{amsfonts}       
\usepackage{nicefrac}       
\usepackage{microtype}      
\usepackage[table]{xcolor}
\definecolor{cellHighlight}{RGB}{235,245,255}
\usepackage{algorithm}
\usepackage{algpseudocode}
\usepackage{color, soul}
\usepackage[most,breakable]{tcolorbox}

\usepackage[labelfont=bf]{caption}
\usepackage{enumitem}
\usepackage{silence}
\usepackage{sectsty}
\usepackage{titlesec}
\titlespacing{\paragraph}{0pt}{0.5em}{0.6em}

\usepackage{fancyhdr}
\AtBeginDocument{\setlength{\headheight}{1cm}}

\definecolor{BMBDarkBlue}{HTML}{315EFE}
\definecolor{BMBLightBlue}{HTML}{00D3ED}
\usepackage[colorlinks, linkcolor=BMBDarkBlue, anchorcolor=BMBDarkBlue, citecolor=BMBDarkBlue, urlcolor=BMBDarkBlue]{hyperref}

\sectionfont{\color{BMBDarkBlue}\fontfamily{zi4}\selectfont}
\subsectionfont{\color{BMBDarkBlue}\fontfamily{zi4}\selectfont}
\subsubsectionfont{\color{BMBDarkBlue}\fontfamily{zi4}\selectfont}

\newtcolorbox{mytheorem}{
  colback=gray!5,       
  colframe=gray!80,     
  boxrule=0.5pt,        
  arc=4pt,              
  left=4pt,             
  right=4pt,            
  top=4pt,              
  bottom=4pt,           
}

\newtcolorbox{promptbox}[1]{
  enhanced jigsaw,
  width=\textwidth,
  colback=BMBDarkBlue!4,
  colframe=BMBDarkBlue!35,
  left=3mm,
  right=3mm,
  top=2mm,
  bottom=2mm,
  boxrule=0.5pt,
  arc=1mm,
  breakable,
  fonttitle=\bfseries,
  coltitle=black,
  title=#1,
  before skip=2mm,
  after skip=2mm
}

\newcommand{\fancyheadname}{\textit{\textbf{UltraData-Code}}}

\title{UltraData-Code: From Raw Repositories to Algorithmically Dense and Task-Oriented Code Data}

\author{
Chengying Tu\textsuperscript{\rm 1,2},
Hengyu Zhao\textsuperscript{\rm 2},
Shuaikang Xue\textsuperscript{\rm 2},
Zhongming Qu\textsuperscript{\rm 2},
Jihao Zhou\textsuperscript{\rm 2}, \\
\bf Xinle Lin\textsuperscript{\rm 2},
Junshao Guo\textsuperscript{\rm 2},
Zixuan Fu\textsuperscript{\rm 3},
Qiang Ma\textsuperscript{\rm 3},
Jie Zhou\textsuperscript{\rm 2},
Chaojun Xiao\textsuperscript{\rm 3}$^{\ddagger}$, \\
\bf 
Hongfei Yan\textsuperscript{\rm 1}$^{\ddagger}$,
Yudong Wang\textsuperscript{\rm 3}$^{\dagger\ddagger}$,
Xu Han\textsuperscript{\rm 3},
Zhiyuan Liu\textsuperscript{\rm 3},
Maosong Sun\textsuperscript{\rm 3}
\\
\textsuperscript{\rm 1}Peking University ~~~~
\textsuperscript{\rm 2}ModelBest Inc. ~~~~
\textsuperscript{\rm 3}Tsinghua University\\
\texttt{tuchengying@stu.pku.edu.cn} \quad\quad
\texttt{yudongwang@tsinghua.edu.cn}
}
\date{}

\newcommand\blfootnote[1]{%
\begingroup
\renewcommand\thefootnote{}\footnote{#1}%
\addtocounter{footnote}{-1}%
\endgroup
}

\begin{document}

\maketitle
\blfootnote{$\ddagger$ Corresponding authors.}
\blfootnote{$\dagger$ Project leaders.}

\thispagestyle{fancy}

\input{sections/0-Abstract}
\input{sections/1-Introduction}
\input{sections/2-Methodology}
\input{sections/3-Experiments}
\input{sections/4-Related-Work}
\input{sections/5-Conclusion}
\input{sections/6-Limitations}

\newpage

\bibliographystyle{citation}
\bibliography{citation}

\clearpage
\appendix
\onecolumn
\input{sections/apx-A-L1-Filtering-Details}
\input{sections/apx-B-L2-Selection-Details}
\input{sections/apx-C-L3-Synthesis-Details}
\input{sections/apx-D-Multilingual-Evaluation-Details}

\end{document}

%% file: sections/0-Abstract.tex
\begin{abstract}
Code generation has become a core capability of large language models (LLMs), and code data is central to the pre-training process that develops it. As obtaining public code of sufficient quality becomes more difficult, further progress cannot rely on corpus expansion alone. Tiered data management instead applies increasingly targeted processing to smaller data pools, producing data states suited to different training objectives. Following this perspective, we present \textbf{\textit{UltraData-Code}} as four connected data states spanning repository archival at L0, standardized natural code at L1, algorithmic selection at L2, and task-oriented synthesis at L3. L0 preserves repository data and provenance for reprocessing, while L1 applies scalable filtering, normalization, and near-deduplication for broad pre-training. L2 selects high-quality algorithmically relevant files using language-adaptive relevance signals and quality constraints, yielding approximately 400B tokens of \textbf{\textit{UltraData-Code-L2}} across 11 programming languages for focused continual pre-training. L3 converts selected implementations into structured programming exercises, yielding approximately 150B tokens of \textbf{\textit{UltraData-Code-L3}} across the same 11 languages for task-oriented mid-training or supervised fine-tuning (SFT).
Under controlled 10B-token continual pre-training of a 1B model, L2 improves pass@1 over L1 by \textbf{7.80} percentage points (\textit{pp}) on EvalPlus and \textbf{5.13\textit{pp}} on MultiPL-E. Replacing half of the L2 training tokens with L3 yields a further gain of \textbf{8.42\textit{pp}} and \textbf{8.07\textit{pp}} on the two benchmarks, respectively. Both gains increase when the training budget scales to 100B tokens. We have released our datasets and classifiers to support further research on LLMs.
\end{abstract}

%% file: sections/1-Introduction.tex
\vspace{-2mm}
\section{Introduction}
\label{sec:introduction}
\vspace{-2mm}

Code generation has become a core capability of foundation models, driving the development of specialized code models~\citep{chen2021evaluating,nijkamp2022codegen,roziere2023code,guo2024deepseek,zhu2024deepseek,song2024alchemistcoder}. Accordingly, constructing and curating code corpora has become a central research focus~\citep{li2023starcoder,huang2025opencoder,seed2025seed}. Their quality, scale, and composition influence models' exposure to programming language structures, program patterns, and problem-solving knowledge during training~\citep{li2024datacomp,grattafiori2024llama,allal2025smollm2,hu2024minicpm,minicpmteam2025minicpm4ultraefficientllmsend}. As obtaining public code of sufficient quality becomes more difficult, further progress cannot rely on corpus expansion alone. Tiered data management instead applies increasingly targeted processing to smaller data pools, producing data states suited to different training objectives. We therefore study how explicit data transitions affect code generation performance under a fixed model, evaluation protocol, and token budget.

\begin{figure}[t]
    \centering
    \includegraphics[width=\textwidth]{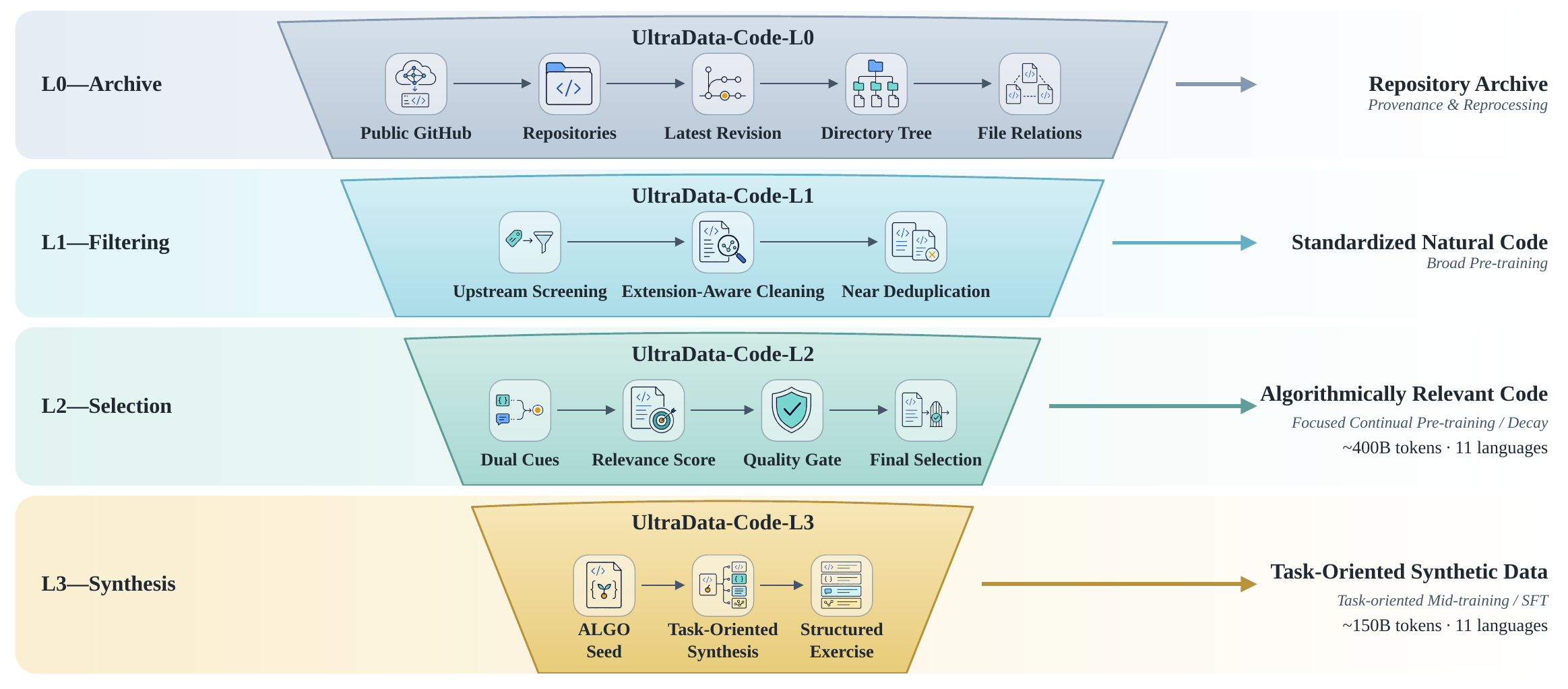}
    \caption{UltraData-Code progresses from repository archival (L0) through standardized natural code (L1) and algorithmic relevance selection (L2) to task-oriented synthesis (L3). L0 supports provenance and reprocessing. L1 supports broad pre-training. L2 supports focused continual pre-training, while L3 supports task-oriented mid-training or SFT.}
    \label{fig:ultradata-code-overview}
\end{figure}

Following the tiered data management perspective~\citep{wang2026datasciencetechnologyagi}, UltraData-Code organizes code data into four connected and separately usable states, as shown in Figure~\ref{fig:ultradata-code-overview}. This organization preserves broad coverage at lower levels while concentrating model-based selection and synthesis on smaller data pools of greater value. L0 archives repository data and provenance for reprocessing. L1 follows the conventional filtering, normalization, and deduplication practices used in existing code corpora~\citep{kocetkov2022stack,lozhkov2024starcoder,huang2025opencoder} to standardize natural code for broad pre-training and supply candidates for L2. L2 introduces an automated and language-adaptive framework that selects high-quality algorithmically relevant code for focused continual pre-training and supplies seeds for L3. L3 grounds task-oriented synthesis in algorithmic files and generates aligned task, analysis, solution, and test fields for mid-training or SFT.

Our construction yields approximately 400B tokens of \textbf{\textit{UltraData-Code-L2}} and 150B tokens of \textbf{\textit{UltraData-Code-L3}} across 11 programming languages. Under controlled 10B-token continual pre-training from the same 1B model checkpoint, L2 improves pass@1 over its L1 counterpart by 7.80 percentage points (\textit{pp}) on EvalPlus and 5.13\textit{pp} on MultiPL-E. Replacing half of the L2 training tokens with L3 yields further gains of 8.42\textit{pp} and 8.07\textit{pp} on the two benchmarks, respectively. These gains persist and widen as the training budget increases to 100B tokens, demonstrating the value of selecting and composing data according to its processing level.

Our main contributions are as follows.

\begin{itemize}[leftmargin=*]
    \item \textbf{A connected code data construction pipeline.} UltraData-Code links repository archival and standardized natural code with selection and structured synthesis, yielding approximately 400B tokens at L2 and 150B tokens at L3 across 11 programming languages. The datasets have been released to facilitate the development of LLMs.
    \item \textbf{Fine-grained curation of algorithmically relevant code.} We develop a language-adaptive selection framework that combines file role supervision with language-specific heuristic cues to learn algorithmic relevance. The framework reuses precomputed semantic embeddings across role, relevance, and quality models, and applies role and quality constraints during selection.
    \item \textbf{Implementation-grounded, task-oriented synthesis.} We introduce a structured synthesis protocol that converts each algorithmically relevant implementation selected by L2 into a programming exercise, jointly generating a standalone task, analysis, solution, and test candidates from the same source implementation. This transformation preserves the source code's computational intent while adding explicit task and solution supervision for code generation.
\end{itemize}

%% file: sections/2-Methodology.tex

\vspace{-2mm}
\section{Tiered Code Data Management}
\label{sec:method}
\vspace{-2mm}

Tiered management moves beyond corpus expansion alone by matching processing effort and data use to the value of each state. Scalable filtering and deduplication are applied to the broad L1 pool, while more expensive model-based selection and synthesis are concentrated on progressively smaller L2 and L3 pools. The resulting tiers remain separately usable. L0 supports provenance and reprocessing. L1 supports broad pre-training and supplies the candidate pool for downstream selection. L2 supports focused continual pre-training and supplies seeds for synthesis. L3 provides task-oriented data for mid-training or SFT.

\subsection{L0: Repository Archival}
\label{sec:l0-collection}

L0 is the source archival state that defines the repository boundary for all subsequent processing. The repository metadata cutoff is July 2025 and covers approximately 192 million public GitHub repositories, whereas the cutoff for repository downloads is October 2025. For each retained repository, we keep only the latest commit on its designated default branch. The archive preserves repository files, directory structures, and metadata such as repository and owner names, dates, and stars. At the file level, it records the content, path, repository, and retained revision of each file. Thus, L0 fixes the data boundary and retained revision while preserving the repository archive that serves as the input to the filtering and deduplication pipeline in L1.

\subsection{L1: Scalable Filtering and Deduplication}
\label{sec:l1-cleaning}

L1 transforms the L0 repository archive into standardized natural code through three stages. \emph{Upstream screening} establishes broad file eligibility, \emph{extension-aware cleaning} repairs localized noise and applies quality gates for each format, and \emph{near deduplication} reduces repeated content while preserving representative files. These stages operate over two levels of organization. Let $\mathcal{C}$ denote the 80 mutually exclusive language categories and $\mathcal{E}$ the 160 selected extensions. Here, an extension is the routing value used by the implementation and may be a conventional suffix or a special filename such as \texttt{Makefile}. Each $e\in\mathcal{E}$ maps to one parent category under $\pi:\mathcal{E}\rightarrow\mathcal{C}$. Categories organize corpus statistics, while extensions select cleaning profiles and define partitions for near deduplication. The complete mapping between categories and extensions is reported in Table~\ref{tab:l1-category-entry-mapping} in Appendix~\ref{app:l1-category-extension-mapping}.

\paragraph{Upstream screening establishes broad file eligibility.}
\label{sec:l1-upstream-screening}
Given a file record $x$ from L0, upstream screening applies path, metadata, encoding, and content checks. It assigns $x$ to a category in $\mathcal{C}$, records binary status and generated content status, rejects unusable text, and marks oversized files, dotfiles outside the allowlist, and files under excluded directories for soft deletion. This stage retains approximately 7.8 billion file records from approximately 85 million repositories. Appendix~\ref{app:l1-upstream-screening} specifies the eligibility rules and resulting language composition.

\paragraph{Extension-aware cleaning applies quality rules for each format.}
\label{sec:l1-extension-aware-cleaning}
For a screened record $x$, its extension $e\in\mathcal{E}$ selects a cleaning profile with repair operators and quality gates appropriate to its format. Extensions with similar failure modes share profiles specialized for source code, web files, assembly, configuration files, and documentation. For example, the \emph{General/Core} profile applies license, PII, secret, and Base64 rewriting before generated code, minification, keyword density, and numeric density checks. The \emph{Web scripts} profile removes source map references and long SVG paths and applies a JavaScript keyword gate. The cleaning profiles and their operations are detailed in Appendix~\ref{app:l1-extension-aware-cleaning}.

Each profile normalizes, rewrites, and then validates its input. For extension $e$, let $R_e$ denote its normalization and rewriting operators, and let $Q_e$ denote its quality gate. The pipeline computes $z=R_e(x)$ and evaluates $Q_e(z)$. A passing record is \texttt{kept} when $z=x$ and \texttt{modified} otherwise. A failing record is marked as \texttt{filtered}. Records in the \texttt{kept} and \texttt{modified} states proceed to the next near deduplication stage.

\paragraph{Near deduplication removes repetition within extension partitions.}
\label{sec:l1-deduplication}
Each extension $e\in\mathcal{E}$ defines an independent partition for near deduplication. Before MinHash construction, the pipeline physically removes records marked for soft deletion during upstream screening. It computes MinHash signatures for the remaining records and retrieves candidate pairs through LSH~\citep{broder1998min,manku2007detecting}. A pair forms an edge when the number of shared bands reaches the threshold configured for extension $e$. Connected components define near-duplicate groups. Files within each component are ranked by repository stars, repository creation time, token length, and a stable identifier. The pipeline retains approximately 50\% of the files in each component, rounding up when the component size is odd, while retaining all files with no detected near duplicates.

Extension-aware cleaning produces 7.11B records. Before MinHash construction, the pipeline removes approximately 472M records marked for soft deletion during upstream screening. Graph construction excludes roughly 72 thousand records without identifiers. Applying the retention rule within each connected component then yields 3.98B files. These operations reduce the cleaning output by 44.05\%. Appendix~\ref{app:l1-near-deduplication} provides the complete configuration and corpus accounting.

\subsection{L2: Algorithmic Relevance Selection}
\label{sec:l2-selection}

Existing code selection methods typically use heuristic rules~\citep{guo2024deepseek} or LLM judgments of educational value~\citep{gunasekar2023textbooks,allal2025smollm2} and code quality~\citep{seed2025seed,fujii2025rewriting}. Heuristic rules are bounded by predefined patterns, whereas LLM-based criteria often emphasize educational or presentation-level properties with limited attention to algorithmic logic. Since algorithmic code explicitly embodies computational and problem-solving logic, incorporating such code into training data can help models develop stronger code generation capabilities. OpenCoder~\citep{huang2025opencoder} has incorporated a small algorithmic code corpus, whose files often concentrate program logic locally and rely less on surrounding files. Using keyword-based rules to identify such files, it shows that including them in the training data can improve code generation. However, these rules can miss algorithmic logic without explicit keyword cues. The problem becomes harder across programming languages because relevant libraries, idioms, and repository conventions differ. A central unresolved question is therefore how to select algorithmically relevant code across programming languages while supporting efficient selector adaptation at corpus scale.

\begin{figure}[t]
    \centering
    \includegraphics[width=0.88\textwidth]{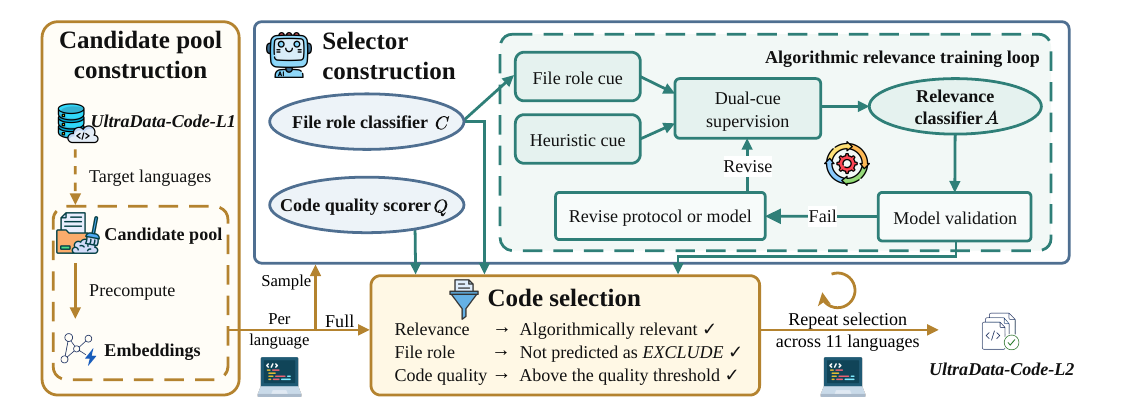}
    \caption{Overview of UltraData-Code-L2 selection framework. We precompute an embedding for every file. For each programming language, we then construct a language-specific selector comprising a file role classifier, an algorithmic relevance classifier, and a code quality scorer. The green dashed box expands the algorithmic relevance training loop: file role and heuristic cues form dual-cue supervision for training the relevance classifier, while validation failures trigger protocol revision. The file role and code quality models use analogous automated training loops. Full-corpus selection combines the language-specific relevance threshold with file role and quality constraints to produce the corresponding language-specific subset.}
    \label{fig:l2_selection_overview}
\end{figure}

At L2, we introduce an automated and language-adaptive framework for selecting algorithmically relevant code to improve the code generation capabilities of foundation models, as illustrated in Figure~\ref{fig:l2_selection_overview}. We first formalize file-level algorithmic relevance and define a shared taxonomy for modeling file roles. We then construct a selector that combines a relevance score learned from dual-cue supervision with file role and quality constraints. Finally, we use a controller to automate cross-language adaptation of selection protocols and coordinate model training, corpus scoring, and full corpus selection.

\subsubsection{Problem Formulation}
\label{sec:l2-overview}

For each programming language $\ell$, the selector evaluates every file in the corresponding language subset of UltraData-Code-L1. We call a file algorithmically relevant when it contains explicit computation or problem-solving logic that is interpretable from its local context. Although files devoted to algorithm implementation often concentrate such logic, files with other primary responsibilities may also contain it. Our goal is to instantiate a language-specific selection rule $\Phi_\ell(x)\in\{0,1\}$, where $\Phi_\ell(x)=1$ indicates that file $x$ is retained. Across programming languages, the selection rules share the same definition of algorithmic relevance. Each rule is instantiated using evidence specific to language $\ell$, allowing it to recognize language-dependent expressions of algorithmic relevance without changing the selection objective.

\subsubsection{File Role Classification}
\label{sec:l2-supervision}

\begin{table}[t]
    \centering
    \caption{Code file role taxonomy based on each file’s primary responsibility. The taxonomy is shared across programming languages, while its annotation criteria are instantiated with language-specific evidence. EXCLUDE takes precedence, and TOOL serves as the fallback role among eligible files.}
    \label{tab:l2_file_role_taxonomy}
    \vspace{1mm}
    \begin{tabular}{@{}p{0.09\columnwidth}p{0.88\columnwidth}@{}}
        \toprule
        \textbf{Category} & \textbf{Operational criterion} \\
        \midrule
        ALGO & Implements algorithms or core data structures for abstract computational problems. \\
        WEB & Handles network requests and responses or defines services, endpoints, and remote interfaces. \\
        TOOL & Provides executable utilities, automation, examples, reusable components, or interfaces. \\
        DATA & Transforms, analyzes, simulates, or learns from data beyond simple I/O or parsing. \\
        TEST & Verifies the behavior or correctness of other code through checks, fixtures, mocks, or test support. \\
        CONFIG & Declares settings, constants, schemas, or metadata without substantive executable logic. \\
        \midrule
        EXCLUDE & Contains generated or repetitive artifacts or is unavailable, empty, stub, or comment-only. \\
        \bottomrule
    \end{tabular}
\end{table}

We define the code file role taxonomy in Table~\ref{tab:l2_file_role_taxonomy} to describe each file's primary responsibility. It identifies algorithm-focused files through ALGO and unsuitable training content through EXCLUDE, while retaining other roles that may still contain useful problem-solving logic. To construct role supervision, we sample files from the language-specific subset of UltraData-Code-L1 and apply a language-specific annotation protocol comprising the classification prompt and structured output rules. Each language contributes 50,000 annotated files, which keeps the supervision scale comparable across languages. The taxonomy and output schema remain shared, while each protocol adapts the recognition evidence and instructions to the target language.
Under this protocol, each sampled file receives exactly one role label from an LLM annotator. For a sampled file $x$, we denote this supervision label by $\tilde c(x)$. To extend these sample-level labels to all files in the corresponding language-specific subset, we use Qwen3-Embedding-0.6B~\citep{zhang2025qwen3} as the embedding model $E$, which represents each file as $\mathbf e(x)=E(x)$. We train a seven-way multilayer perceptron (MLP) classifier $C_\ell$ on the labeled pairs $\bigl(\mathbf e(x),\tilde c(x)\bigr)$. After training, its prediction for any file in that language subset is
\begin{equation}
\hat c_\ell(x)=C_\ell\bigl(\mathbf e(x)\bigr).
\end{equation}
Thus, $\tilde c(x)$ denotes a supervision label available only for the annotated sample, whereas $\hat c_\ell(x)$ denotes the classifier prediction available at corpus scale. These LLM-derived labels are operational supervision targets. This distillation design limits LLM annotation to sampled files, while precomputing the embeddings allows 
\begin{wrapfigure}[11]{r}{0.48\textwidth}
    \centering
    \includegraphics[width=0.46\textwidth]{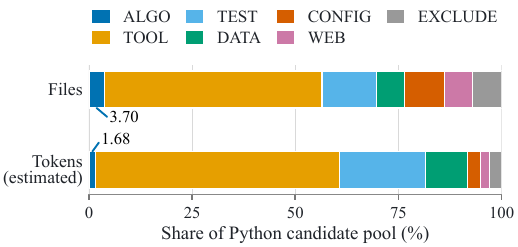}
    \caption{File and estimated token shares of the Python candidate pool by the roles predicted by $C_{\mathrm{py}}$.}
    \label{fig:l2-file-role-share}
\end{wrapfigure}
the same file representations to be reused by subsequent selection models.
Appendix~\ref{app:l2-language-protocol} provides the selection protocol and annotation details.

Applying $C_{\mathrm{py}}$ to UltraData-Code-L1-py reveals a concentrated role distribution by both file count and estimated token volume. As shown in Figure~\ref{fig:l2-file-role-share}, TOOL accounts for 52.80\% of files and an estimated 59.03\% of tokens, whereas ALGO accounts for only 3.70\% of files and 1.68\% of the estimated token volume. Files centered on algorithm implementation therefore form a precise but sparse anchor, motivating a complementary cue that can identify algorithmic evidence beyond the ALGO role. The controlled role probes and further analysis of relevance scoring and dual-cue coverage across file roles are detailed in Appendix~\ref{app:l2-selection-analysis}.

\subsubsection{Selector Construction}
\label{sec:l2-selector}

To broaden supervision beyond files centered on algorithm implementation, we combine two signals. The file role cue marks annotated ALGO files, while the heuristic cue searches for language-specific functions, libraries, and control-flow patterns inside other file roles~\citep{huang2025opencoder}. For each annotated file $x$, let $h_\ell(x,\tilde c(x))\in\{0,1\}$ indicate whether the language-specific, role-conditioned heuristic protocol identifies algorithmic evidence in the code under its role label $\tilde c(x)$. The dual-cue pseudo-label is
\begin{equation}
\tilde y_{A,\ell}(x)=
\mathbf 1
\left[
\tilde c(x)=\mathrm{ALGO}
\ \lor\
h_\ell(x,\tilde c(x))=1
\right].
\end{equation}
An ALGO label or a positive heuristic match yields a positive pseudo-label. Files matched by neither cue form the negative set. The role cue anchors supervision in ALGO, while the heuristic cue extends it to other roles. Role identity therefore supplies supervision without becoming the final selection rule. Appendix~\ref{app:l2-dual-cue-construction} details cue construction, sampling, and balancing.

The algorithmic relevance classifier $A_\ell$ uses these pseudo-labels as targets and combines code semantics with lightweight repository context. We derive a fixed-dimensional hashed feature $\mathbf p(x)$ from the repository-relative path and concatenate it with the contextual code embedding using a fixed weight $\lambda_p=0.4$:
\begin{equation}
\mathbf v(x)=\left[\mathbf e(x);\lambda_p\mathbf p(x)\right].
\end{equation}
The code embedding captures semantics across variations in identifiers, libraries, and surface forms, while the path feature supplies compact structural cues about the file's repository role. An MLP maps the combined representation to the continuous relevance score
\begin{equation}
s_{A,\ell}(x)=A_\ell\bigl(\mathbf v(x)\bigr)\in[0,1].
\end{equation}
The classifier learns a relevance score from semantic and structural evidence, allowing the supervision cues to generalize beyond exact heuristic matches.

Furthermore, code presentation quality provides a separate selection signal. Following SwallowCode-v2~\citep{fujii2025rewriting}, we apply the Seed-Coder~\citep{seed2025seed} quality prompt to sampled files and train the embedding-based code-quality scorer $Q_\ell$. It reuses $\mathbf e(x)$ and produces the expected quality score $s_{Q,\ell}(x)=Q_\ell\bigl(\mathbf e(x)\bigr)\in[0,10]$, with larger values indicating better code quality.

We deterministically assign each file to exactly one data split. The training split fits model parameters, while the validation split selects the training epoch and model configuration. For $A_\ell$, a separate 10\% calibration split selects the relevance threshold after the model is frozen. Appendix~\ref{app:l2-selection-models} provides the architectures, path featurization, and optimization settings for all three selection models. For the frozen models, the final selector is
\begin{equation}
\label{eq:l2-final-selector}
\begin{aligned}
\Phi_\ell(x)
={}&
\mathbf 1
\left[
\hat c_\ell(x)\neq\mathrm{EXCLUDE}
\right]
\mathbf 1
\left[
s_{A,\ell}(x)\geq\tau_{A,\ell}
\right] \\
&\cdot
\mathbf 1
\left[
s_{Q,\ell}(x)\geq\tau_Q
\right].
\end{aligned}
\end{equation}
Here, the relevance threshold $\tau_{A,\ell}$ is the minimum relevance score required to retain a file in language $\ell$. The quality threshold $\tau_Q$ is shared across languages. Relevance is the primary criterion, while the EXCLUDE and quality conditions remove unsuitable files.

We choose $\tau_{A,\ell}$ on the calibration split after the relevance model is frozen. A threshold is feasible when retained files achieve at least 0.85 precision against the dual-cue pseudo-labels and include at least 80\% of ALGO files in the split. Feasible thresholds are ordered by $F_{0.5}$, overall precision, ALGO recall, and threshold value. This procedure selects $\tau_{A,\mathrm{py}}=0.85$ for Python. Appendix~\ref{app:l2-threshold-selection} presents the full threshold selection rule.

\subsubsection{Automated Cross-Language Adaptation}
\label{sec:l2-multilingual}

Algorithmic relevance has a common meaning across programming languages, but the evidence that reveals it depends on local libraries, frameworks, syntax, and idioms. Every language uses the same seven file roles, dual-cue rule, model interfaces, data splits, and threshold criteria. The language-specific protocol supplies the concrete recognition signals.

For each target language, the controller refines the annotation instructions and constructs a heuristic module from representative libraries and code patterns. Protocol selection uses only a disjoint pilot split. By default, the controller revises the protocol for at most five rounds, then freezes a passing version or the strongest completed version with valid taxonomy outputs. The annotation, validation, and calibration splits remain disjoint throughout this process. This adapts evidence recognition while preserving the shared relevance definition.

After training, the frozen models score the specific language subset from L1, and the language-specific relevance threshold completes each selector. Every file is encoded once, and its precomputed representation is reused by the file role classifier, algorithmic relevance classifier, and code quality scorer. Protocol revisions can therefore update sampled supervision and lightweight model heads without recomputing corpus embeddings. Applying the selectors to the corresponding 11 language subsets of UltraData-Code-L1 produces UltraData-Code-L2.
L2 retains approximately 12.23\% of files from L1 across 11 programming languages and contains approximately 400B tokens. It is enriched in files that implement algorithms while retaining algorithmically relevant code from other file roles. The statistics of L2 are reported in Appendix~\ref{app:l2-data-statistics}.

\paragraph{Transfer to an independent source corpus.}
To test whether the L2 selection framework can transfer to an independent source corpus, we construct a variant from Stack-v3-train~\citep{lozhkov2026stackv3}. We conduct a preliminary quality check and cleaning of the dataset. The cleaning stage filters URL-only records, overly short records, and records containing garbled or malformed text. We then apply the same L2 selectors to the cleaned dataset and produce \textbf{\textit{UltraData-Stack-v3-L2}}. It covers the same 11 programming languages as UltraData-Code-L2, and detailed statistics by language are reported in Appendix~\ref{app:l2-ultra-stack-v3}.

\subsection{L3: Task-Oriented Synthesis}
\label{sec:l3-synthetic}

Existing efforts to synthesize code training corpora have primarily followed two directions. Textbook-style synthesis constructs explicit programming tasks together with solution reasoning and reference implementations, emphasizing problem-solving supervision~\citep{gunasekar2023textbooks}. Code rewriting instead transforms existing implementations by improving their interfaces, documentation, dependencies, and overall presentation quality, thereby preserving realistic code context~\citep{fujii2025rewriting}. These directions motivate a data construction objective that connects explicit programming tasks with realistic implementation structure.

\begin{figure}[ht]
    \centering
    \includegraphics[width=\textwidth]{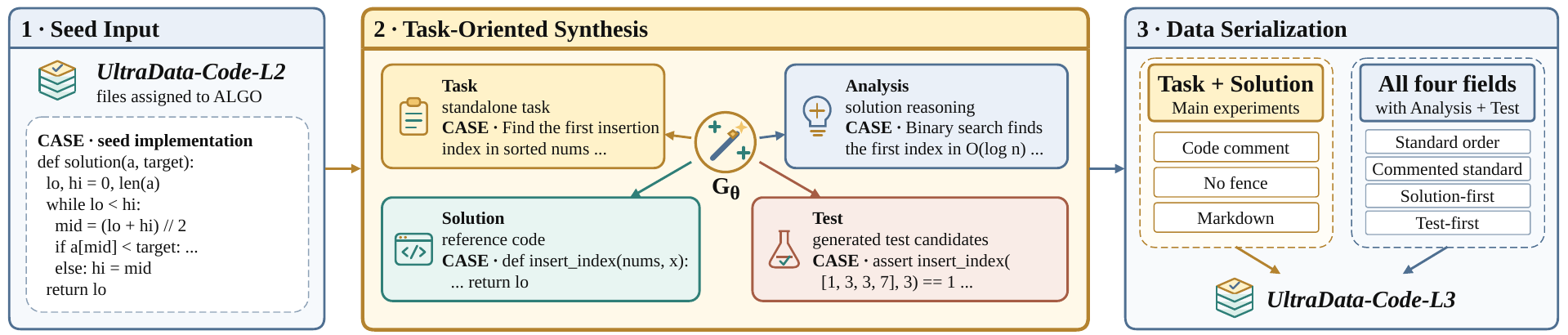}
    \caption{Overview of UltraData-Code-L3 synthesis. Files assigned to ALGO at L2 provide seed implementations. A single call to $G_\theta$ converts each seed into four aligned fields: a standalone task, an analysis of the solution reasoning, a standalone reference solution, and generated test candidates. Serialization produces either task and solution sequences for the main experiments, using code comment, no fence, or Markdown formatting, or sequences containing all four fields in standard, commented, solution-first, or test-first order.}
    \label{fig:l3-synthesis}
\end{figure}

At L3, we realize this objective through an implementation-grounded protocol that transforms selected ALGO implementations into self-contained programming exercises. We use DeepSeek-v4-flash in no-think mode as the synthesis model $G_\theta$. As shown in Figure~\ref{fig:l3-synthesis}, the workflow separates structured exercise generation from training sequence serialization. The following subsections describe how each seed is converted into a structured record and how that record is formatted for continual pre-training.

\subsubsection{Generation from ALGO Seeds}
\label{sec:l3-task-code-construction}

Let $\mathcal X=\{x_i\mid i=1,\ldots,N\}$ be the set of ALGO seeds. For each seed $x_i$, the synthesis model produces one structured programming exercise in a single generation call. A shared prompt schema, instantiated with syntax and interface requirements for the target language, asks the model to use $x_i$ as inspiration for a standalone programming task and generate four explicitly delimited fields:
\begin{equation}
\label{eq:l3-structured-synthesis}
z_i=G_{\theta}(x_i)=\left(t_i,a_i,s_i,u_i\right).
\end{equation}
Here $t_i$ is a standalone task specification, $a_i$ explains the algorithm, edge cases, and complexity, $s_i$ is a self-contained reference solution, and $u_i$ contains test candidates for the specified behavior. The task must be independent of the source snippet, while the other fields must solve, explain, and test that task. The parser retains records containing all four explicitly delimited fields. Appendix~\ref{app:l3-synthesis-details} provides the generation prompt and structural parsing details.

\subsubsection{Data Serialization}
\label{sec:l3-serialization}

Each L3 record contains four complementary fields, so the training signal depends on how these fields are selected, ordered, and presented during continual pre-training. For example, CodeAlchemy~\citep{gupta2026codealchemy} organizes synthesized code into question answering, enhancement, and dialogue forms, illustrating how synthesis form can target different downstream code capabilities. A single serialization format repeatedly presents the same field order and boundary pattern. We therefore separate content generation from serialization and construct multiple training views from each structured record.

Let $\mathcal F=\{\mathrm{task},\mathrm{analysis},\mathrm{solution},\mathrm{test}\}$ denote the field types in $z_i=(t_i,a_i,s_i,u_i)$. For a selected subset $S\subseteq\mathcal F$, the projection $\pi_S(z_i)$ retains the corresponding fields. A presentation format $f$ determines their order, delimiters, and code representation. The final training sample is
\begin{equation}
\label{eq:l3-serialization}
y_i^{(S,f)}=\operatorname{Serialize}_{f}\!\left(\pi_S(z_i)\right).
\end{equation}
UltraData-Code-L3 uses $S_{\mathrm{TS}}=\{\mathrm{task},\mathrm{solution}\}$ with code comment, no fence, and Markdown formats. No fence directly concatenates the task description and solution without code fences. The Markdown format uses code fences annotated with the target language. Code comment places the task context before the implementation using a comment or string literal representation. UltraData-Code-L3 (w/ Analysis\&Test) uses $S=\mathcal F$ with the standard order $(t_i,a_i,s_i,u_i)$, a commented version of that order, a solution-first order, or a test-first order.

These choices reflect our broader principle that synthetic records should be serialized in diverse forms rather than tied to a single recurring template. The released records preserve all four fields, and the reported size of UltraData-Code-L3 counts their combined content. Appendix~\ref{app:l3-serialization-templates} provides the complete templates and allocation details.

%% file: sections/3-Experiments.tex
\vspace{-2mm}
\section{Experiments}
\vspace{-2mm}
\label{sec:experiments}

We evaluate the effects of UltraData-Code-L2 and UltraData-Code-L3 through controlled continual pre-training. We first use the Python subsets for focused, small-scale validation and then extend the evaluation to a multilingual setting covering 11 programming languages. In both settings, we measure the gains from L2 selection and test whether adding L3 data provides further improvements under matched token budgets.

\subsection{Experimental Setup}
\label{sec:experimental-setup}

\paragraph{Training details.}
All continual pre-training experiments start from the same 1B checkpoint, obtained after approximately 1.3T tokens of pre-training, and use the same optimization configuration across data settings. Detailed model configurations are provided in Table~\ref{tab:pretraining-architecture}, where \textit{Params.}, \textit{Vocab.}, $d_m$, $d_{ff}$, $d_h$, $n_q$, $n_{kv}$, and $L$ denote the total number of non-embedding parameters, vocabulary size, model hidden dimension, feedforward intermediate dimension, attention head dimension, number of query heads, number of key-value heads, and number of layers, respectively. It uses a SentencePiece tokenizer with 73,448 tokens, a micro batch size of 10, and a global batch size of 1280. The peak and minimum learning rates are $7.5\times10^{-4}$ and $3.75\times10^{-5}$, respectively. We use a WSD schedule with exponential decay during the final phase of training and FP8 GEMM. A one-layer multi-token prediction head provides an auxiliary cross-entropy loss weighted by 0.1. Each 10B-token run trains for approximately three hours on 32 NVIDIA H100 GPUs.

\begin{table}[ht]
    \centering
    \caption{Model configuration for the experimental model.}
    \label{tab:pretraining-architecture}
    \vspace{1mm}
    \begin{tabular}{llcccccc}
        \toprule
        \textit{Params.} & \textit{Vocab.} & $d_m$ & $d_{ff}$ & $d_h$ & $n_q$ & $n_{kv}$ & $L$ \\
        \midrule
        1,321,322,496 & 73,448 & 2,048 & 6,144 & 128 & 16 & 2 & 28 \\
        \bottomrule
    \end{tabular}
\end{table}

\paragraph{Data preparation and comparison controls.}
To isolate the effect of data construction under the shared training configuration above, we standardize corpus preparation and sampling across all runs. Our experiments cover four corpora from the UltraData-Code pipeline: UltraData-Code-L1, UltraData-Code-L2, UltraData-Code-L3, and UltraData-Stack-v3-L2. Before training, every corpus undergoes the same cleaning, sampling, and benchmark-decontamination pipeline, beginning with exact 13-gram matching. For L1--L2 comparisons, the multilingual UltraData-Code-L1 control is restricted to the same 11 languages as UltraData-Code-L2 and sampled to match its language-level token mixture; Python runs use the corresponding Python subsets. For L3 comparisons, each synthetic corpus is mixed with UltraData-Code-L2 at a 1:1 token ratio, or with UltraData-Code-L2-py in the Python setting. The reported size of UltraData-Code-L3 includes all four generated fields, while the main experiments use only the task and solution fields. Table~\ref{tab:exp-dataset-overview} summarizes the language coverage and approximate token volume of each experimental and baseline corpus.

\paragraph{Evaluation.}
We evaluate the code generation capabilities of models after continual pre-training using OpenCompass~\citep{2023opencompass}. For Python, we evaluate EvalPlus~\citep{liu2023your} and report pass@1 on HumanEval~\citep{chen2021evaluating}, HumanEval+, MBPP~\citep{austin2021program}, and MBPP+. For evaluation across multiple programming languages, we use MultiPL-E~\citep{cassano2023multipl} and report pass@1. All models use identical task prompts and inference configurations with greedy decoding.

\begin{table}[t]
    \centering
    \caption{Language coverage and approximate token counts of the experimental and baseline datasets.}
    \label{tab:exp-dataset-overview}
    \vspace{1mm}
    \small
    \setlength{\tabcolsep}{5pt}
    \begin{tabular}{llrr}
        \toprule
        \textbf{Level} & \textbf{Dataset} & \textbf{\# Languages} & \textbf{\# Tokens} \\
        \midrule
        \multirow{3}{*}{L1}
        & Stack-v2 & 619 & $\sim$900B \\
        & RefineCode & 607 & $\sim$960B \\
        & Stack-v3-train & 713 & $\sim$4.9T \\
        \midrule
        \multirow{4}{*}{L2}
        & Stack-Edu & 15 & $\sim$125B \\
        & Swallow-v2-high & 1 & $\sim$50B \\
        &\cellcolor{cellHighlight}UltraData-Stack-v3-L2 & \cellcolor{cellHighlight}11 & \cellcolor{cellHighlight}$\sim$240B \\
        & \cellcolor{cellHighlight}UltraData-Code-L2 & \cellcolor{cellHighlight}11 & \cellcolor{cellHighlight}$\sim$400B \\
        \midrule
        \multirow{3}{*}{L3}
        & CodeAlchemy & 15 & $\sim$976B \\
        & SwallowCode-v2 & 1 & $\sim$50B \\
        & \cellcolor{cellHighlight}UltraData-Code-L3 & \cellcolor{cellHighlight}11 & \cellcolor{cellHighlight}$\sim$150B \\
        \bottomrule
    \end{tabular}
\end{table}

\paragraph{Baselines.}
We organize the comparison datasets by the data management level at which they are constructed: L1 applies preliminary filtering and deduplication, L2 performs model-based selection, and L3 performs refinement or synthesis~\citep{wang2026datasciencetechnologyagi}.

\begin{itemize}[leftmargin=*]
    \item \textbf{L1:} The Stack v2 (Stack-v2)~\citep{lozhkov2024starcoder} is built from Software Heritage with basic cleaning, deduplication, and sensitive information processing, and provides a broad untargeted reference. RefineCode~\citep{huang2025opencoder} applies more than 130 general and language-specific heuristic rules to GitHub code and code-related Web data. Stack-v3-train~\citep{lozhkov2026stackv3} is a large filtered corpus collected from public repositories, spanning 713 programming and markup languages, and providing a recent broad-coverage reference for evaluating selection beyond an earlier corpus source.
    \item \textbf{L2:} Stack-Edu~\citep{allal2025smollm2} trains language-specific classifiers to select code with educational value. Swallow-v2-high is the high-quality Python subset derived from SwallowCode-v2~\citep{fujii2025rewriting} using the Seed-Coder~\citep{seed2025seed} quality prompt.
    \item \textbf{L3:} CodeAlchemy~\citep{gupta2026codealchemy} is a large-scale repository-grounded synthesis corpus that transforms source code into various formats, with its code-qa, code-enhance, and code-dialogue variants used as synthetic baselines. SwallowCode-v2~\citep{fujii2025rewriting} is a rewriting-based corpus that targets code presentation quality, readability, completeness, and independent executability, with its high-quality Python subset Swallow-v2-high used as a quality-focused synthetic baseline.
\end{itemize}

All baseline comparisons use the full multilingual corpus or its corresponding Python subset where available, with matched training budgets and identical optimization and evaluation settings. \textit{In the subsequent experimental analysis, the suffix ``-py'' identifies a Python-only subset.}

\subsection{Main Results}
\label{sec:experimental-results}

\subsubsection{Python Results}
\label{sec:l2-python-results}

Table~\ref{tab:python-all-results} compares the L1--L3 data states under the same 10B-token continual pre-training budget. We use the Python setting to separate three questions: whether algorithmic relevance selection improves natural code training, whether this improvement transfers across source corpora, and whether structured synthesis provides an additional gain beyond the selected natural code foundation.

\begin{table}[t]
    \centering
    \caption{Python code generation results on EvalPlus after continual pre-training on 10B tokens. L3 data are mixed with UltraData-Code-L2-py at a 1:1 token ratio.}
    \label{tab:python-all-results}
    \vspace{1mm}
    \small
    \setlength{\tabcolsep}{2.4pt}
    \begin{tabular}{clccccc}
        \toprule
        \textbf{Level} & \textbf{Dataset} & \textbf{HumanEval} & \textbf{HumanEval+} & \textbf{MBPP} & \textbf{MBPP+} & \textbf{AVG.} \\
        \midrule
        \multirow{4}{*}{\textbf{L1}} & Stack-v2 & 10.98 & 9.15 & 28.31 & 22.75 & 17.80 \\
        & RefineCode & 17.68 & 12.80 & 39.68 & 30.42 & 25.15 \\
        & Stack-v3-train-py & 18.90 & 17.68 & 43.92 & 35.45 & 28.99 \\
        & \cellcolor{cellHighlight}UltraData-Code-L1-py & \cellcolor{cellHighlight}18.29 & \cellcolor{cellHighlight}16.46 & \cellcolor{cellHighlight}36.51 & \cellcolor{cellHighlight}29.63 & \cellcolor{cellHighlight}25.22 \\
        \midrule
        \multirow{5}{*}{\textbf{L2}} & Stack-Edu & 18.90 & 17.07 & 39.42 & 32.80 & 27.05 \\
        & Swallow-v2-high & 23.78 & 20.12 & 42.33 & 35.71 & 30.49 \\
        & Stack-Edu-py & 25.00 & 19.51 & 51.85 & 42.33 & 34.67 \\
        & \cellcolor{cellHighlight}UltraData-Stack-v3-L2-py & \cellcolor{cellHighlight}37.80 & \cellcolor{cellHighlight}34.15 & \cellcolor{cellHighlight}\textbf{52.12} & \cellcolor{cellHighlight}\textbf{43.39} & \cellcolor{cellHighlight}41.87 \\
        & \cellcolor{cellHighlight}UltraData-Code-L2-py & \cellcolor{cellHighlight}\textbf{43.29} & \cellcolor{cellHighlight}\textbf{39.02} & \cellcolor{cellHighlight}49.21 & \cellcolor{cellHighlight}41.01 & \cellcolor{cellHighlight}\textbf{43.13} \\
        \midrule
        \multirow{5}{*}{\textbf{L3}} & CodeAlchemy-code-qa-py & 29.88 & 24.39 & 49.21 & 41.53 & 36.25 \\
        & CodeAlchemy-code-enhance-py & 35.37 & 29.88 & 50.79 & 42.33 & 39.59 \\
        & CodeAlchemy-code-dialogue-py & 34.15 & 29.88 & 47.09 & 38.36 & 37.37 \\
        & SwallowCode-v2 & 40.85 & 34.15 & 51.59 & 41.27 & 41.97 \\
        & \cellcolor{cellHighlight}UltraData-Code-L3-py & \cellcolor{cellHighlight}\textbf{46.34} & \cellcolor{cellHighlight}\textbf{39.63} & \cellcolor{cellHighlight}\textbf{53.97} & \cellcolor{cellHighlight}\textbf{45.77} & \cellcolor{cellHighlight}\textbf{46.43} \\
        \bottomrule
    \end{tabular}
\end{table}

\begin{figure}[ht]
    \centering
    \includegraphics[width=\textwidth]{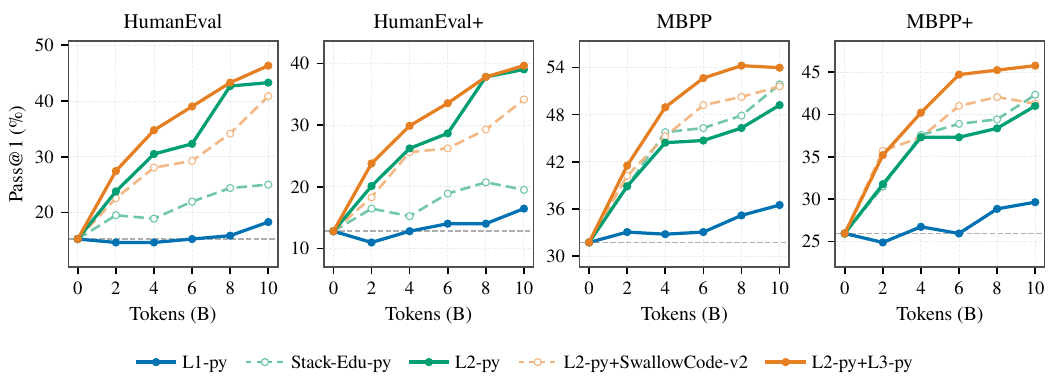}
    \caption{EvalPlus training dynamics for datasets of different levels. Solid lines denote our datasets, while dashed lines denote baselines.}
    \label{fig:python-training-dynamics}
\end{figure}

\paragraph{Algorithmic relevance selection substantially improves natural code training.}
UltraData-Code-L2-py improves by 17.91\textit{pp} on EvalPlus over its direct L1 source, UltraData-Code-L1-py, under the same 10B-token budget, achieving the best overall performance at L2. It also exceeds the strongest L2 baseline, Stack-Edu-py, by 8.46\textit{pp} in average pass@1.

\paragraph{The L2 improvement is not tied to a single source corpus.}
The independently sourced UltraData-Stack-v3-L2-py improves by 12.88\textit{pp} over its Stack-v3-train-py source and also records the highest MBPP and MBPP+ scores. Together with the direct UltraData-Code comparison, this result indicates that our algorithmic relevance selection transfers across source corpora.

\paragraph{Structured L3 synthesis provides an additional gain beyond L2.}
The equal-token mixture of UltraData-Code-L2-py and UltraData-Code-L3-py reaches an average pass@1 of 46.43, which is 3.30\textit{pp} above L2-only training and 4.46\textit{pp} above the strongest L3 mixture baseline. The improvement is consistent across all four Python benchmarks. This result indicates that task-oriented synthesis complements the selected natural code foundation, providing more structured programming exercises.

\paragraph{The tiered data advantage emerges early and persists throughout training.}
As shown in Figure~\ref{fig:python-training-dynamics}, UltraData-Code-L2-py separates from UltraData-Code-L1-py early in training, while the L2--L3 mixture becomes the strongest setting from 4B tokens onward and remains ahead through 10B tokens on all four EvalPlus subsets. The persistence of these margins indicates that the observed gains are not confined to a transient final checkpoint effect.

\subsubsection{Multilingual Results}
\label{sec:l2-multilingual-results}

Table~\ref{tab:l2-multilingual-main-results} evaluates L1--L3 data states across 11 programming languages under the same 10B-token continual pre-training budget. We use this setting to test whether L2 selection extends across languages and whether L3 synthesis adds a multilingual gain beyond L2. Language-level HumanEval and MBPP results are provided in Appendix~\ref{app:multipl-e-results}.

\begin{table}[t]
    \centering
    \caption{Code generation results across multiple programming languages after continual pre-training on 10B tokens. L3 data are mixed with UltraData-Code-L2 at a 1:1 token ratio.}
    \label{tab:l2-multilingual-main-results}
    \vspace{1mm}
    \small
    \setlength{\tabcolsep}{2.4pt}
    \begin{tabular}{clcccccccc}
        \toprule
        & & \multicolumn{5}{c}{\textbf{EvalPlus}} & \multicolumn{3}{c}{\textbf{MultiPL-E}} \\
        \cmidrule(lr){3-7}\cmidrule(lr){8-10}
        \textbf{Level} & \textbf{Dataset} & \textbf{HE} & \textbf{HE+} & \textbf{MBPP} & \textbf{MBPP+} & \textbf{AVG.} & \textbf{HE} & \textbf{MBPP} & \textbf{AVG.} \\
        \midrule
        \multirow{4}{*}{\textbf{L1}} & Stack-v2 & 10.98 & 9.15 & 28.31 & 22.75 & 17.80 & 7.77 & 14.27 & 11.02 \\
        & RefineCode & 17.68 & 12.80 & 39.68 & 30.42 & 25.15 & 12.37 & 21.27 & 16.82 \\
        & Stack-v3-train & 20.73 & 18.29 & 35.19 & 27.25 & 25.37 & 12.57 & 22.02 & 17.30 \\
        & \cellcolor{cellHighlight}UltraData-Code-L1 & \cellcolor{cellHighlight}17.68 & \cellcolor{cellHighlight}14.63 & \cellcolor{cellHighlight}34.66 & \cellcolor{cellHighlight}27.51 & \cellcolor{cellHighlight}23.62 & \cellcolor{cellHighlight}10.80 & \cellcolor{cellHighlight}19.69 & \cellcolor{cellHighlight}15.25 \\
        \midrule
        \multirow{3}{*}{\textbf{L2}} & Stack-Edu & 18.90 & 17.07 & 39.42 & 32.80 & 27.05 & 11.78 & 22.89 & 17.33 \\
        & \cellcolor{cellHighlight}UltraData-Stack-v3-L2 & \cellcolor{cellHighlight}23.78 & \cellcolor{cellHighlight}19.51 & \cellcolor{cellHighlight}\textbf{42.59} & \cellcolor{cellHighlight}34.13 & \cellcolor{cellHighlight}30.00 & \cellcolor{cellHighlight}18.16 & \cellcolor{cellHighlight}\textbf{27.34} & \cellcolor{cellHighlight}\textbf{22.75} \\
        & \cellcolor{cellHighlight}UltraData-Code-L2 & \cellcolor{cellHighlight}\textbf{26.83} & \cellcolor{cellHighlight}\textbf{21.34} & \cellcolor{cellHighlight}\textbf{42.59} & \cellcolor{cellHighlight}\textbf{34.92} & \cellcolor{cellHighlight}\textbf{31.42} & \cellcolor{cellHighlight}\textbf{14.56} & \cellcolor{cellHighlight}26.20 & \cellcolor{cellHighlight}20.38 \\
        \midrule
        \multirow{5}{*}{\textbf{L3}} & CodeAlchemy-code-qa & 28.66 & 25.61 & 45.77 & 37.04 & 34.27 & 16.29 & 25.02 & 20.65 \\
        & CodeAlchemy-code-enhance & 24.39 & 21.34 & 40.74 & 34.39 & 30.22 & 16.08 & 25.16 & 20.62 \\
        & CodeAlchemy-code-dialogue & 20.12 & 17.68 & 41.01 & 34.92 & 28.43 & 11.92 & 21.88 & 16.90 \\
        & \cellcolor{cellHighlight}UltraData-Code-L3 (w/ Analysis\&Test) & \cellcolor{cellHighlight}31.10 & \cellcolor{cellHighlight}24.39 & \cellcolor{cellHighlight}\textbf{52.65} & \cellcolor{cellHighlight}41.80 & \cellcolor{cellHighlight}37.49 & \cellcolor{cellHighlight}21.42 & \cellcolor{cellHighlight}29.15 & \cellcolor{cellHighlight}25.28 \\
        & \cellcolor{cellHighlight}UltraData-Code-L3 & \cellcolor{cellHighlight}\textbf{33.54} & \cellcolor{cellHighlight}\textbf{29.27} & \cellcolor{cellHighlight}52.38 & \cellcolor{cellHighlight}\textbf{44.18} & \cellcolor{cellHighlight}\textbf{39.84} & \cellcolor{cellHighlight}\textbf{24.94} & \cellcolor{cellHighlight}\textbf{31.97} & \cellcolor{cellHighlight}\textbf{28.45} \\
        \bottomrule
    \end{tabular}
\end{table}

\paragraph{Language-adaptive selection improves multilingual code generation.}
UltraData-Code-L2 reaches 31.42 on the EvalPlus average and 20.38 on the MultiPL-E average, improving by 7.80\textit{pp} and 5.13\textit{pp} over its UltraData-Code-L1 source, respectively. It also exceeds Stack-Edu by 4.37\textit{pp} on EvalPlus and 3.05\textit{pp} on MultiPL-E. As the L2 selector adapts its relevance cues and decision thresholds to each programming language, these gains extend the selection effect beyond the focused Python setting to multiple languages.

\paragraph{Implementation-grounded synthesis adds a second gain beyond selection.}
Both L3 variants outperform the strongest synthetic baseline, CodeAlchemy-code-qa. With task and solution only, UltraData-Code-L3 reaches 39.84 on EvalPlus and 28.45 on MultiPL-E, adding 8.42\textit{pp} and 8.07\textit{pp} over L2-only training and exceeding CodeAlchemy-code-qa by 5.57\textit{pp} and 7.80\textit{pp}, respectively. UltraData-Code-L3 (w/ Analysis\&Test), which retains all four generated fields, reaches 37.49 and 25.28, still above CodeAlchemy-code-qa. The task-and-solution variant performs better under the 10B-token budget because it concentrates more tokens on the task-solution relationship, while the four-field variant allocates part of the same budget to analysis and test. Both results show that turning selected implementations into structured exercises adds supervision beyond selected natural code alone.

\begin{figure}[htbp]
    \centering
    \includegraphics[width=\textwidth]{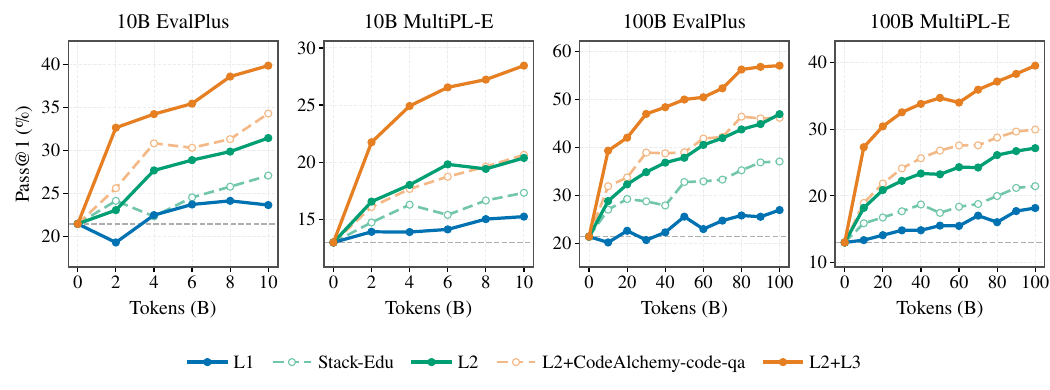}
    \caption{Multilingual training dynamics at 10B and 100B tokens. The first two panels show the 10B-token experiments, while the last two show dedicated scaling runs of 100B tokens.}
    \label{fig:multilingual-training-scaling-dynamics}
\end{figure}

\paragraph{The two gains persist through training and grow with scale.}
In the first two panels of Figure~\ref{fig:multilingual-training-scaling-dynamics}, L2 separates from L1 early, and the L2--L3 mixture leads on both benchmark averages from 2B tokens onward through 10B tokens. In the last two panels of Figure~\ref{fig:multilingual-training-scaling-dynamics}, the 100B-token trajectories amplify the same ordering, with final scores shown in Table~\ref{tab:l2-scaling-final}. At the final checkpoint, L2 exceeds L1 by 19.98\textit{pp} on EvalPlus and 8.99\textit{pp} on MultiPL-E, while replacing half of L2 with L3 adds another 10.11\textit{pp} and 12.39\textit{pp} over L2-only training. The mixture reaches 57.06 and 39.54 mean pass@1.

\begin{table}[t]
    \centering
    \caption{Final multilingual results after continual pre-training on 100B tokens. L3 settings mix the indicated synthetic corpus with UltraData-Code-L2 at a 1:1 token ratio.}
    \label{tab:l2-scaling-final}
    \vspace{1mm}
    \setlength{\tabcolsep}{8pt}
    \begin{tabular}{lcc}
        \toprule
        \textbf{Dataset} & \textbf{EvalPlus AVG.} & \textbf{MultiPL-E AVG.} \\
        \midrule
        \cellcolor{cellHighlight}UltraData-Code-L1 & \cellcolor{cellHighlight}26.97 & \cellcolor{cellHighlight}18.16 \\
        Stack-Edu & 37.09 & 21.44 \\
        \cellcolor{cellHighlight}UltraData-Code-L2 & \cellcolor{cellHighlight}46.95 & \cellcolor{cellHighlight}27.15 \\
        L2 + CodeAlchemy-code-qa & 46.20 & 29.94 \\
        \cellcolor{cellHighlight}L2 + UltraData-Code-L3 & \cellcolor{cellHighlight}\textbf{57.06} & \cellcolor{cellHighlight}\textbf{39.54} \\
        \bottomrule
    \end{tabular}
\end{table}

\subsection{Ablation Study}
\label{sec:ablations}

For L2, we examine the signals used for L2 selection. For L3, we compare different synthesis methods and examine the effects of modifying the natural code of the mixture during the experiment.

\subsubsection{L2 Ablations}
\label{sec:l2-selection-ablation}

To isolate the contributions of the L2 selection signals, we compare five Python settings under the same 10B-token continual pre-training setup. \texttt{UltraData-Code-L2-py} combines the algorithmic relevance score with \texttt{EXCLUDE} removal and code quality constraints. As shown in Table~\ref{tab:l2-selection-ablation}, \texttt{Relevance-only} uses only the algorithmic score selection, \texttt{Relevance + EXCLUDE} additionally removes files predicted as EXCLUDE, and \texttt{Relevance + Quality} instead adds only the code quality threshold. \texttt{Quality-only} uses only the code quality selection.

\begin{table}[htbp]
    \centering
    \caption{Ablation of Python data selection strategies on EvalPlus after continual pre-training on 10B tokens.}
    \label{tab:l2-selection-ablation}
    \vspace{1mm}
    \setlength{\tabcolsep}{2.4pt}
    \begin{tabular}{lccccc}
        \toprule
        \textbf{Setting} & \textbf{HumanEval} & \textbf{HumanEval+} & \textbf{MBPP} & \textbf{MBPP+} & \textbf{AVG.} \\
        \midrule
        \cellcolor{cellHighlight}UltraData-Code-L2-py & \cellcolor{cellHighlight}\textbf{43.29} & \cellcolor{cellHighlight}\textbf{39.02} & \cellcolor{cellHighlight}\textbf{49.21} & \cellcolor{cellHighlight}\textbf{41.01} & \cellcolor{cellHighlight}\textbf{43.13} \\
        \midrule
        Relevance-only & 31.10 & 28.05 & 48.68 & 40.74 & 37.14 \\
        Relevance + EXCLUDE & 32.93 & 31.10 & 48.41 & 40.74 & 38.30 \\
        Relevance + Quality & 30.49 & 28.05 & 48.68 & 39.95 & 36.79 \\
        Quality-only & 18.29 & 16.46 & 34.13 & 27.25 & 24.03 \\
        \bottomrule
    \end{tabular}
\end{table}

\paragraph{Algorithmic relevance is the primary driver of downstream gains.} \texttt{Relevance-only} outperforms \texttt{Quality-only} by 13.11\textit{pp} in average pass@1. The complete strategy adds 5.99\textit{pp} over \texttt{Relevance-only}, showing that the auxiliary constraints enhance the relevance signal.

\paragraph{The auxiliary constraints are most effective when applied together.} \texttt{Relevance-only}, \texttt{Relevance + EXCLUDE}, and \texttt{Relevance + Quality} achieve comparable average pass@1 scores, whereas applying the quality threshold after removing EXCLUDE files raises average pass@1 from 38.30 to 43.13. Moreover, as detailed in Appendix~\ref{app:l2-selection-models}, 92.26\% of the Python data already satisfies the shared quality threshold, confirming that quality acts as an auxiliary veto that removes a limited low-score tail after unsuitable files have been excluded.

\subsubsection{L3 Ablations}
\label{sec:l3-content-ablation}

We compare four synthesis controls with \texttt{UltraData-Code-L3-py} under the same 10B-token budget in Table~\ref{tab:l3-content-objectives}. Every setting uses 5B tokens from \texttt{UltraData-Code-L2-py} and 5B synthetic tokens. \texttt{Textbook Synthesis} and \texttt{Code Rewriting} instantiate the two individual routes~\citep{gunasekar2023textbooks,fujii2025rewriting}. \texttt{Corpus-Level Mixture} allocates the synthetic budget across examples independently produced by these routes. \texttt{Assembling} is derived from the same two routes at the example level: for each shared ALGO seed, it pairs the task produced by \texttt{Textbook Synthesis} with the implementation produced by \texttt{Code Rewriting}.

\begin{table}[ht]
    \vspace{-2mm}
    \centering
    \caption{Ablation of Python data synthesis methods on EvalPlus after continual pre-training on 10B tokens.}
    \label{tab:l3-content-objectives}
    \vspace{1mm}
    \setlength{\tabcolsep}{2.4pt}
    \begin{tabular}{lccccc}
        \toprule
        \textbf{Strategy} & \textbf{HumanEval} & \textbf{HumanEval+} & \textbf{MBPP} & \textbf{MBPP+} & \textbf{AVG.} \\
        \midrule
        \cellcolor{cellHighlight}UltraData-Code-L3-py & \cellcolor{cellHighlight}46.34 & \cellcolor{cellHighlight}39.63 & \cellcolor{cellHighlight}53.97 & \cellcolor{cellHighlight}45.77 & \cellcolor{cellHighlight}\textbf{46.43} \\
        \midrule
        Textbook Synthesis & 40.85 & 34.76 & \textbf{54.50} & 45.50 & 43.90 \\
        Code Rewriting & \textbf{48.78} & \textbf{42.07} & 50.53 & 42.33 & 45.93 \\
        Corpus-Level Mixture & 43.29 & 39.02 & 53.17 & 44.71 & 45.05 \\
        Assembling & 45.12 & 39.63 & 53.97 & \textbf{46.83} & 46.39 \\
        \bottomrule
    \end{tabular}
    \vspace{-2mm}
\end{table}

\paragraph{Code rewriting gives the strongest single-route result.}
\texttt{Code Rewriting} is the strongest individual route at 45.93 average pass@1, compared with 43.90 for \texttt{Textbook Synthesis}. It leads on HumanEval and HumanEval+, while \texttt{Textbook Synthesis} leads on MBPP. \texttt{Corpus-Level Mixture} reaches 45.05 and remains below \texttt{Code Rewriting}, indicating that combining independently generated examples does not fully integrate the strengths of the two routes.

\paragraph{Pairing both views at the example level is more effective than corpus-level mixing.}
\texttt{Assembling} reaches 46.39 average pass@1, exceeding \texttt{Corpus-Level Mixture} by 1.34\textit{pp} and achieving the highest MBPP+ score. Because \texttt{Assembling} pairs the textbook-style task and rewritten implementation derived from the same ALGO seed, this comparison isolates the benefit of aligning the two views within each training example rather than placing them in separate corpus-level examples. \texttt{UltraData-Code-L3-py} achieves the highest average pass@1 at 46.43, showing that its unified implementation-grounded synthesis matches or exceeds this explicitly assembled control.

\paragraph{The natural code base shapes the gains from L3 synthesis.}
We compare the same synthetic sources when paired with UltraData-Code-L1-py in Table~\ref{tab:l3-l1-mixture}. To match the main L3 Python experiments, each mixture uses a 1:1 token ratio under the same 10B-token budget. In Table~\ref{tab:python-all-results}, only the mixture containing UltraData-Code-L3-py improves over the UltraData-Code-L2-py baseline. By contrast, when the synthetic sources are paired with UltraData-Code-L1-py, all of them improve over L1-only training. This comparison shows that the benefit of synthetic data also depends on the natural code corpus used in the mixture.

\begin{table}[ht]
    \vspace{-2mm}
    \centering
    \caption{Python EvalPlus results for L1:L3 mixtures. The L1-only baseline uses 10B L1 tokens.}
    \label{tab:l3-l1-mixture}
    \vspace{1mm}
    \small
    \setlength{\tabcolsep}{2.4pt}
    \begin{tabular}{lccccc}
        \toprule
        \textbf{Synthetic data} & \textbf{HumanEval} & \textbf{HumanEval+} & \textbf{MBPP} & \textbf{MBPP+} & \textbf{AVG.} \\
        \midrule
        \cellcolor{cellHighlight}UltraData-Code-L1-py & \cellcolor{cellHighlight}18.29 & \cellcolor{cellHighlight}16.46 & \cellcolor{cellHighlight}36.51 & \cellcolor{cellHighlight}29.63 & \cellcolor{cellHighlight}25.22 \\
        \midrule
        SwallowCode-v2 & 25.00 & 21.95 & 45.77 & 36.24 & 32.24 \\
        CodeAlchemy-code-qa-py & 27.44 & 24.39 & 39.42 & 33.33 & 31.15 \\
        CodeAlchemy-code-enhance-py & 22.56 & 19.51 & 42.59 & 34.92 & 29.90 \\
        CodeAlchemy-code-dialogue-py & 19.51 & 15.85 & 40.21 & 32.54 & 27.03 \\
        \cellcolor{cellHighlight}UltraData-Code-L3-py & \cellcolor{cellHighlight}\textbf{37.80} & \cellcolor{cellHighlight}\textbf{32.32} & \cellcolor{cellHighlight}\textbf{51.32} & \cellcolor{cellHighlight}\textbf{42.06} & \cellcolor{cellHighlight}\textbf{40.88} \\
        \bottomrule
    \end{tabular}
    \vspace{-2mm}
\end{table}

%% file: sections/4-Related-Work.tex
\vspace{-2mm}
\section{Related Work}
\vspace{-2mm}

\subsection{Code Corpus Construction}

Large code corpora built from public repositories provide the scale and language coverage needed for code model pre-training. The Stack~\citep{kocetkov2022stack} establishes a permissively licensed source code corpus and makes source licensing an explicit part of corpus design. StarCoder~\citep{li2023starcoder} and Code Llama~\citep{roziere2023code} demonstrate how such corpora can support open code foundation models. The Stack v2~\citep{lozhkov2024starcoder} expands repository coverage through Software Heritage and strengthened source provenance. The Stack v3~\citep{lozhkov2026stackv3} continues this trajectory toward larger and more comprehensively documented repository archives.

Turning repository archives into training corpora requires a separate standardization process. Existing code corpora combine language identification, license and metadata screening, removal of generated or sensitive content, and near-deduplication~\citep{kocetkov2022stack,lozhkov2024starcoder}. Deduplication can improve model quality~\citep{lee2022deduplicating}, while corpus cleaning must also account for duplication across code datasets because overlap between training and evaluation data can compromise evaluation reliability~\citep{lopez2024inter}. These practices establish a progression from preserving repository sources to producing standardized natural code. Tiered data management formalizes this progression by assigning distinct purposes to successive data states~\citep{wang2026datasciencetechnologyagi}.

\subsection{Code Selection}

Code selection methods differ in both the properties they target and the scale at which they operate. For code corpora, DeepSeek-Coder~\citep{guo2024deepseek} uses heuristic filtering to remove low-quality content, while Stack-Edu selects data for educational value~\citep{allal2025smollm2}. Other approaches use learned assessments of code quality or task alignment. Seed-Coder~\citep{seed2025seed} uses a model's judgments to curate its own training data, and Arctic-SnowCoder~\citep{wei2024arcticsnowcoder} trains a code-quality annotator and studies alignment with downstream task distributions. Large-scale text corpora also provide evidence for systematic filtering and verification, as demonstrated by FineWeb and Ultra-FineWeb~\citep{penedo2024fineweb,wang2025ultrafinewebefficientdatafiltering}. At a broader level, importance resampling and influence estimation by a model assess the utility of individual examples~\citep{xie2023data,yu2024mates}, mixture optimization adjusts the contribution of different data sources~\citep{xie2023doremi}, and DataComp-LM~\citep{li2024datacomp} compares candidate corpora under controlled training settings. Collaborative selection and expanding reasoning data further extend this line of work~\citep{bai2025efficient,zhang2026expanding}.

A separate question is whether selection directly captures computational or task relevance. FINDR~\citep{zhang2025findr} combines hashed n-gram features, code features, and influence priors to identify data aligned with NL2Code tasks. OpenCoder~\citep{huang2025opencoder} uses keyword rules to extract code that implements algorithms and demonstrates its value for code model training. FINDR focuses on alignment with a specified task, whereas OpenCoder relies on explicit lexical evidence of algorithmic intent. Keyword rules can miss computational logic expressed through data processing, web services, engineering tools, and tests, and their cues must be adapted across programming languages. Our L2 selection instead learns a continuous relevance score from semantic representations supervised by file role and heuristic cues defined for each language. It then combines this score with separate constraints for unsuitable roles and code quality.

\subsection{Code Synthesis}

Code synthesis has different scale requirements across stages of LLM training, such as pre-training and SFT. These differences lead to distinct synthesis objectives and methods. Program synthesis formalizes the mapping from natural language specifications to executable programs~\citep{austin2021program,tu2025surveyllmmidtraining}, while SFT methods such as Magicoder's OSS-Instruct use public code snippets to construct instruction and solution pairs~\citep{wei2024magicoder}. For pre-training, the objective shifts toward corpus-scale synthesis that can support sustained exposure during model training. Textbook synthesis constructs structured explanations, programming tasks, and reference implementations~\citep{gunasekar2023textbooks}. Code rewriting transforms existing implementations to improve their presentation and independent execution while retaining realistic code context~\citep{fujii2025rewriting}. Other work expands the source and generation process through comment augmentation~\citep{song2024code}, multi-source integration~\citep{song2024alchemistcoder}, synthesis grounded in repository context~\citep{gupta2026codealchemy}, and agentic generation and refinement~\citep{mi2026data,deng2026dataevolver,kulikov2026autodata}. Building on these complementary directions, our L3 transforms selected algorithmic implementations into structured programming tasks at corpus scale.

%% file: sections/5-Conclusion.tex
\vspace{-2mm}
\section{Conclusion}
\label{sec:conclusion}
\vspace{-2mm}

In this paper, we present \textbf{\textit{UltraData-Code}}, a tiered framework that transforms repository archives into four separately usable data states. Its central principle is to move beyond corpus expansion alone by applying increasingly targeted processing to smaller data pools and aligning each resulting state with an appropriate use. L0 preserves repository data and provenance for reprocessing, while L1 standardizes natural code for broad pre-training and supplies candidates for L2. At L2, a language-adaptive selector combines file role and heuristic supervision with role and quality constraints to produce algorithmically relevant code for focused continual pre-training and seeds for L3. At L3, task-oriented synthesis converts selected L2 implementations into aligned task, analysis, solution, and test fields for mid-training or SFT. Together, these stages produce approximately 400B tokens of L2 data and 150B tokens of L3 data across 11 programming languages.

Controlled continual pre-training isolates the contribution of these data transitions. With a budget of 10B tokens, L2 improves average pass@1 over L1 by 7.80\textit{pp} on EvalPlus and 5.13\textit{pp} on MultiPL-E and also exceeds Stack-Edu. Replacing half of the L2 training tokens with L3 adds a further 8.42\textit{pp} and 8.07\textit{pp} on the two benchmarks and surpasses the strongest evaluated synthetic baseline. At a scale of 100B tokens, the improvements increase to 19.98\textit{pp} and 8.99\textit{pp} from L1 to L2, followed by another 10.11\textit{pp} and 12.39\textit{pp} after incorporating L3. These results show that algorithmic relevance selection and task-oriented synthesis provide complementary mechanisms for converting large-scale repository code into more effective code corpora. This work aims to provide systematic insights and practical resources for code data construction, offering valuable support for future research on LLM training and contributing to the further development of LLMs.

%% file: sections/6-Limitations.tex
\vspace{-2mm}
\section{Limitations and Future Directions}
\label{sec:limitations}
\vspace{-2mm}

Some key limitations of our work are as follows. Due to time and resource constraints, the current L2 algorithmic relevance selection only covers 11 programming languages, and we hope to extend it to other programming languages. In the multilingual comparison between L1 and L2, we match the token mixture across languages in the two datasets to study the data transition under a controlled composition. This matched language mixture provides a controlled basis for evaluating the effect of L2 selection, while the multilingual scores of baseline corpora may be affected by their different language distributions. Experiments with alternative language mixtures and their effects on downstream performance remain to be explored. At L3, we introduce multiple serialization formats to reduce the risk that synthetic data collapses into a fixed surface pattern. The effects of their proportions and of additional formats on data diversity remain to be explored. The L2 and L3 mixture uses a fixed 1:1 token ratio to maintain a fair comparison, and we plan to study how this ratio changes when synthetic data is incorporated into a practical continual pre-training pipeline. These efforts further refine our tiered code data management framework and provide better-quality data support for training LLMs.

%% file: sections/apx-A-L1-Filtering-Details.tex
\section{L1 Filtering Details}
\label{app:l1-filtering-details}

In this appendix, we report the mapping between categories and extensions, screening rules and corpus statistics, cleaning profiles, and the configuration for near-deduplication in L1.

\subsection{Category and Extension Mapping}
\label{app:l1-category-extension-mapping}

The implementation routes 160 selected extensions into 80 mutually exclusive language categories. An extension may be a conventional suffix or a special filename such as \texttt{Makefile} or \texttt{Dockerfile}. Each extension belongs to one category, while a category may contain multiple extensions. The complete mapping is shown in Table~\ref{tab:l1-category-entry-mapping}.

\begingroup
\scriptsize
\setlength{\tabcolsep}{3pt}
\renewcommand{\arraystretch}{1.08}
\begin{longtable}{p{0.115\textwidth}r p{0.285\textwidth}p{0.115\textwidth}r p{0.285\textwidth}}
    \caption{Complete mapping from the 80 language categories to the 160 selected extensions. }
    \label{tab:l1-category-entry-mapping} \\
    \toprule
    \textbf{Category} & \textbf{\#} & \textbf{Extensions} & \textbf{Category} & \textbf{\#} & \textbf{Extensions} \\
    \midrule
    \endfirsthead
    \multicolumn{6}{c}{\tablename\ \thetable\ continued} \\
    \toprule
    \textbf{Category} & \textbf{\#} & \textbf{Extensions} & \textbf{Category} & \textbf{\#} & \textbf{Extensions} \\
    \midrule
    \endhead
    \midrule
    \multicolumn{6}{r}{Continued on next page} \\
    \endfoot
    \bottomrule
    \endlastfoot
    \texttt{C} & 2 & \texttt{c, h} & \texttt{Dart} & 1 & \texttt{dart} \\
    \texttt{Python} & 2 & \texttt{py, pyi} & \texttt{Dhall} & 1 & \texttt{dhall} \\
    \texttt{JavaScript} & 4 & \texttt{js, jsx, mjs, cjs} & \texttt{Elixir} & 2 & \texttt{ex, exs} \\
    \texttt{PHP} & 2 & \texttt{php, phtml} & \texttt{EmacsLisp} & 1 & \texttt{el} \\
    \texttt{CPP} & 7 & \texttt{cpp, hpp, cc, inl, cxx, hh, hxx} & \texttt{Erlang} & 2 & \texttt{erl, hrl} \\
    \texttt{CSS} & 4 & \texttt{css, scss, less, sass} & \texttt{FSharp} & 1 & \texttt{fs} \\
    \texttt{Java} & 1 & \texttt{java} & \texttt{GraphQL} & 2 & \texttt{graphql, gql} \\
    \texttt{CSharp} & 2 & \texttt{cs, csx} & \texttt{Groovy} & 2 & \texttt{gradle, groovy} \\
    \texttt{HTML} & 2 & \texttt{html, htm} & \texttt{HCL} & 1 & \texttt{hcl} \\
    \texttt{Markdown} & 2 & \texttt{md, markdown} & \texttt{Idris} & 1 & \texttt{idr} \\
    \texttt{TypeScript} & 2 & \texttt{ts, tsx} & \texttt{Isabelle} & 1 & \texttt{thy} \\
    \texttt{Assembly} & 2 & \texttt{s, asm} & \texttt{JSON} & 3 & \texttt{json, jsonl, geojson} \\
    \texttt{Go} & 1 & \texttt{go} & \texttt{Kotlin} & 2 & \texttt{kt, kts} \\
    \texttt{Ruby} & 3 & \texttt{rb, Gemfile, Rakefile} & \texttt{Lean} & 1 & \texttt{lean} \\
    \texttt{Rust} & 1 & \texttt{rs} & \texttt{Mathematica} & 1 & \texttt{nb} \\
    \texttt{SQL} & 1 & \texttt{sql} & \texttt{Matlab} & 2 & \texttt{m, mat} \\
    \texttt{Shell} & 3 & \texttt{sh, zsh, bash} & \texttt{Nim} & 1 & \texttt{nim} \\
    \texttt{Lua} & 1 & \texttt{lua} & \texttt{Nix} & 1 & \texttt{nix} \\
    \texttt{Swift} & 1 & \texttt{swift} & \texttt{OCaml} & 2 & \texttt{ml, mli} \\
    \texttt{Fortran} & 4 & \texttt{f, f90, for, f77} & \texttt{ObjectiveC} & 1 & \texttt{mm} \\
    \texttt{Perl} & 3 & \texttt{pl, pm, t} & \texttt{Pascal} & 2 & \texttt{pas, dpr} \\
    \texttt{TeX} & 3 & \texttt{tex, cls, sty} & \texttt{Prolog} & 1 & \texttt{pro} \\
    \texttt{Makefile} & 3 & \texttt{Makefile, mk, mak} & \texttt{Protobuf} & 1 & \texttt{proto} \\
    \texttt{R} & 1 & \texttt{r} & \texttt{Puppet} & 1 & \texttt{pp} \\
    \texttt{CMake} & 2 & \texttt{CMakeLists.txt, cmake} & \texttt{Racket} & 1 & \texttt{rkt} \\
    \texttt{Scala} & 2 & \texttt{scala, sc} & \texttt{SageMath} & 1 & \texttt{sage} \\
    \texttt{Batchfile} & 2 & \texttt{cmd, bat} & \texttt{Scheme} & 2 & \texttt{scm, ss} \\
    \texttt{VisualBasic} & 3 & \texttt{vb, bas, vbs} & \texttt{Solidity} & 1 & \texttt{sol} \\
    \texttt{PowerShell} & 3 & \texttt{ps1, psm1, psd1} & \texttt{TOML} & 1 & \texttt{toml} \\
    \texttt{Haskell} & 2 & \texttt{hs, lhs} & \texttt{Terraform} & 2 & \texttt{tf, tfvars} \\
    \texttt{Dockerfile} & 2 & \texttt{Dockerfile, dockerfile} & \texttt{VHDL} & 2 & \texttt{vhd, vhdl} \\
    \texttt{Julia} & 1 & \texttt{jl} & \texttt{Verilog} & 3 & \texttt{sv, svh, vh} \\
    \texttt{Ada} & 3 & \texttt{adb, ads, ada} & \texttt{VimScript} & 1 & \texttt{vim} \\
    \texttt{Agda} & 1 & \texttt{agda} & \texttt{WebAssembly} & 2 & \texttt{wast, wat} \\
    \texttt{CUDA} & 2 & \texttt{cu, cuh} & \texttt{XML} & 5 & \texttt{xml, xsd, xsl, dtd, xslt} \\
    \texttt{Clojure} & 4 & \texttt{clj, edn, cljs, cljc} & \texttt{YAML} & 2 & \texttt{yaml, yml} \\
    \texttt{CommonLisp} & 3 & \texttt{lisp, cl, lsp} & \texttt{Zig} & 1 & \texttt{zig} \\
    \texttt{Coq} & 1 & \texttt{v} & \texttt{DiffPatch} & 2 & \texttt{diff, patch} \\
    \texttt{Crystal} & 1 & \texttt{cr} & \texttt{Documentation} & 8 & \texttt{adoc, asc, asciidoc, org, pod, rest, rst, textile} \\
    \texttt{D} & 1 & \texttt{d} & \texttt{Text} & 2 & \texttt{text, txt} \\
\end{longtable}
\endgroup

\subsection{Upstream Screening}
\label{app:l1-upstream-screening}

Upstream screening first uses path and content signals to assign each file to a language category and record its binary status and generated content status. It then applies inexpensive eligibility checks before content rewriting. Files larger than 3\,MiB, dotfiles outside the allowlist, and files under version control, dependency, cache, editor, build, coverage, vendor, and external directories are marked for soft deletion. Generated content status remains an annotation at this stage. Cleaning profiles with a filter for generated code make a separate rejection decision based on content.

Content validation then evaluates textual usability rather than compilation or semantic correctness. Depending on the check, it rejects unusable files or removes limited encoding corruption before passing eligible files to extension-aware cleaning. Table~\ref{tab:l1-content-validation} summarizes these decisions.

\begin{table}[htbp]
    \centering
    \caption{Content validation rules applied during upstream screening.}
    \label{tab:l1-content-validation}
    \vspace{1mm}
    \small
    \setlength{\tabcolsep}{4pt}
    \renewcommand{\arraystretch}{1.08}
    \begin{tabularx}{\textwidth}{@{}p{0.22\textwidth}X@{}}
        \toprule
        \textbf{Check} & \textbf{Decision rule} \\
        \midrule
        Textual content & Require alphabetic characters to cover at least 25\% of the file. For Motorola 68K assembly and WebAssembly, count letters and digits. \\
        Encoded spans & Reject if any Base64, hexadecimal dump, or Unicode escape match exceeds 1,024 characters, or if these matches collectively cover more than 50\% of the file. \\
        UTF-8 validity & Reject if invalid bytes exceed 25\% of the file; otherwise, remove the invalid byte sequences. \\
        Text and structured data & Reject text, JSON, YAML, YML, OWL, and DOT inputs longer than 512 lines. \\
        HTML visible content & Require at least 100 visible characters and a visible text ratio of at least 20\%. \\
        Extensionless text & Retain only designated names, including \texttt{README}, \texttt{NOTES}, \texttt{TODO}, \texttt{DESC}, and \texttt{CMakeLists}. \\
        \bottomrule
    \end{tabularx}
\end{table}

After these checks, upstream screening retains 7,784,171,682 file records from approximately 85 million repositories. Table~\ref{tab:l1-post-screening-composition} reports the largest language categories, with the remaining categories grouped as \emph{Other}.

\begin{table}[htbp]
    \centering
    \caption{File composition after upstream screening. The largest categories are shown individually, and the remaining categories are grouped as \emph{Other}.}
    \label{tab:l1-post-screening-composition}
    \vspace{1mm}
    \setlength{\tabcolsep}{5pt}
    \begin{tabular}{lrr}
        \toprule
        \textbf{Language category} & \textbf{File records} & \textbf{Share} \\
        \midrule
        C & 1,426,392,319 & 18.32\% \\
        Python & 1,249,237,794 & 16.05\% \\
        JavaScript & 794,220,978 & 10.20\% \\
        PHP & 561,237,256 & 7.21\% \\
        JSON & 466,164,719 & 5.99\% \\
        Java & 456,252,952 & 5.86\% \\
        CSS & 371,608,256 & 4.77\% \\
        TypeScript & 313,290,860 & 4.02\% \\
        C\# & 294,936,677 & 3.79\% \\
        Markdown & 262,766,122 & 3.38\% \\
        XML & 252,156,452 & 3.24\% \\
        C++ & 247,414,383 & 3.18\% \\
        HTML & 246,503,494 & 3.17\% \\
        Other & 841,989,420 & 10.82\% \\
        \midrule
        Total & 7,784,171,682 & 100.00\% \\
        \bottomrule
    \end{tabular}
\end{table}

\subsection{Extension-Aware Cleaning}
\label{app:l1-extension-aware-cleaning}

Each extension selects one cleaning profile, and extensions with similar failure modes arising from their formats may share a profile. Extensions without a dedicated registry rule use \emph{General/Core}. Each profile first applies \emph{mappers} that normalize or rewrite content and then applies \emph{filters} that reject records violating quality conditions. For example, a mapper redacts a secret, whereas a filter for generated code rejects the record. Table~\ref{tab:l1-extension-strategies} groups all 160 extensions by profile and summarizes the operations.

\begin{table}[htbp]
    \centering
    \caption{Extension-aware cleaning profiles. The extensions column reports the number assigned to each profile. All profiles also apply common handling for encoding, whitespace, line endings, and garbled text.}
    \label{tab:l1-extension-strategies}
    \vspace{1mm}
    \footnotesize
    \setlength{\tabcolsep}{4pt}
    \renewcommand{\arraystretch}{1.16}
    \begin{tabularx}{\textwidth}{@{}>{\raggedright\arraybackslash}p{0.20\textwidth}>{\centering\arraybackslash}p{0.10\textwidth}>{\raggedright\arraybackslash}p{0.27\textwidth}>{\raggedright\arraybackslash}X@{}}
        \toprule
        \textbf{Profile} & \textbf{Extensions} & \textbf{Additional mappers} & \textbf{Additional filters and thresholds} \\
        \midrule
        \emph{General/Core} & 55 & License, PII, secret, Base64, and excessive line break rewriting & Minimum 30 characters and 2 lines; generated code; minified code (line $\leq$1,000, average $\leq$100); keyword density $\geq$5\%; numeric density $\leq$50\% \\
        \emph{Web scripts} & 6 & Base64, source map URL, and long SVG rewriting & Minimum 30 characters and 2 lines; minified code; JavaScript keyword density $\geq$3\%; numeric density $\leq$50\% \\
        \emph{HTML and CSS} & 6 & Base64 and long SVG rewriting; source map URL removal for CSS & Minimum 30 characters and 2 lines; minified code; HTML visible text $\geq$100 characters and 20--80\% ratio; CSS selector/declaration structure \\
        \emph{Assembly} & 2 & Excessive line break rewriting & Minimum 30 characters and 2 lines; paragraph length $\leq$1,000; unpunctuated span $\leq$1,500; opcode ratio $\geq$5\%; data section ratio $\leq$80\% \\
        \emph{Markdown} & 2 & License, PII, secret, Base64, and excessive line break rewriting & Minimum 30 characters and 2 lines; generated code; numeric density $\leq$50\% \\
        \emph{Makefile} & 3 & License, PII, secret, Base64, and excessive line break rewriting & Markdown gates plus minified code (line $\leq$1,000, average $\leq$100) \\
        \emph{Broad source languages} & 62 & License, PII, secret, Base64, and excessive line break rewriting & Minimum 30 characters and 2 lines; minified code; numeric density $\leq$50\% \\
        \emph{Data and configuration} & 12 & Base64, excessive line break, and secret rewriting & Minimum 10 characters and 1 line \\
        \emph{Text and documentation} & 12 & Excessive line break rewriting & Garbled text filter only \\
        \midrule
        \textbf{Total} & \textbf{160} & & \\
        \bottomrule
    \end{tabularx}
\end{table}

All profiles apply mappers before filters. Applicable minimum size gates are evaluated before and after rewriting so that cleaning cannot leave empty or fragmentary records. Table~\ref{tab:l1-extension-strategies} reports the remaining operations and thresholds for each format.

Generated code rejection is enabled for the \emph{General/Core}, \emph{Markdown}, and \emph{Makefile} profiles. It scans the first 1,000 characters for markers such as ``auto-generated,'' ``do not edit,'' and ``generated by,'' removing approximately 137 million records. Overall, extension-aware cleaning retains approximately 7.1 billion of the 7.8 billion screened records and reduces logical content volume from roughly 62 to 51\,TiB.

\subsection{Near Deduplication}
\label{app:l1-near-deduplication}

Each extension partition forms an independent comparison pool across repositories. Within a partition, the job lowercases content, removes Unicode diacritics, and normalizes whitespace before tokenization. Records with at least five tokens are represented by shingles of five tokens and 255 64-bit MinHash values arranged as 17 LSH bands with 15 hashes per band.

The MinHash signatures define a similarity graph within each partition. Candidate pairs are retrieved from shared LSH bands, and a pair becomes an edge when the number of shared bands reaches the threshold of one, two, or three configured for that extension. Connected components define the near-duplicate groups.

Files within each component are ranked by repository stars, repository creation time, token length, and a stable identifier. For a component of size \(n\), the retention rule keeps \(\max(1,\lceil 0.5n\rceil)\) files. Singleton components are retained.

Before MinHash construction, the pipeline removes approximately 472 million JSON and XML records marked for soft deletion during upstream screening. Graph construction excludes roughly 72 thousand records without identifiers, and retention within components yields 3.98 billion files. These operations reduce the cleaning output by 44.05\%. Table~\ref{tab:l1-language-end-to-end} reports exact counts and rates for the 11 partitions used by L2.

\begin{table}[htbp]
    \centering
    \caption{End-to-end L1 counts for the 11 extension partitions used as inputs to L2 for each language. Removal rates are computed within each partition.}
    \label{tab:l1-language-end-to-end}
    \vspace{1mm}
    \small
    \setlength{\tabcolsep}{2.0pt}
    \begin{tabular}{lrrrrrrr}
        \toprule
        \textbf{Partition} & \textbf{Clean in} & \textbf{Clean out} & \textbf{Dedup out} & \makecell{\textbf{Clean}\\\textbf{removal}} & \makecell{\textbf{Dedup}\\\textbf{removal}} & \makecell{\textbf{End-to-end}\\\textbf{removal}} & \makecell{\textbf{Min.}\\\textbf{votes}} \\
        \midrule
        Python/py & 1,196,338,645 & 1,054,894,245 & 575,184,540 & 11.82\% & 45.47\% & 51.92\% & 3 \\
        Java/java & 456,252,952 & 441,078,485 & 315,317,503 & 3.33\% & 28.51\% & 30.89\% & 2 \\
        C++/cpp & 134,947,528 & 128,992,393 & 80,822,744 & 4.41\% & 37.34\% & 40.11\% & 2 \\
        C\#/cs & 294,904,542 & 284,531,096 & 180,608,749 & 3.52\% & 36.52\% & 38.76\% & 3 \\
        Go/go & 44,239,024 & 42,580,688 & 28,411,832 & 3.75\% & 33.28\% & 35.78\% & 1 \\
        Ruby/rb & 62,118,529 & 57,040,842 & 35,638,854 & 8.17\% & 37.52\% & 42.63\% & 3 \\
        JavaScript/js & 722,038,812 & 624,489,007 & 384,390,857 & 13.51\% & 38.45\% & 46.76\% & 3 \\
        PHP/php & 557,983,174 & 507,114,980 & 290,654,304 & 9.12\% & 42.68\% & 47.91\% & 3 \\
        R/r & 5,618,906 & 5,066,717 & 4,042,704 & 9.83\% & 20.21\% & 28.05\% & 1 \\
        Rust/rs & 16,164,537 & 15,290,126 & 10,801,809 & 5.41\% & 29.35\% & 33.18\% & 1 \\
        Shell/sh & 31,963,304 & 25,989,609 & 16,298,462 & 18.69\% & 37.29\% & 49.01\% & 2 \\
        \midrule
        \textbf{Total} & \textbf{3,522,569,953} & \textbf{3,187,068,188} & \textbf{1,922,172,358} & \textbf{9.52\%} & \textbf{39.69\%} & \textbf{45.43\%} & -- \\
        \bottomrule
    \end{tabular}
\end{table}

%% file: sections/apx-B-L2-Selection-Details.tex
\section{L2 Selection Details}
\label{app:l2-selection-details}

Our framework constructs a selector for each programming language through four stages: selection protocol adaptation and sampled annotation, dual-cue pseudo-label construction, selection model training and validation, and full-corpus scoring with threshold calibration and selection. Each protocol binds its annotation instructions, heuristic evidence, model configuration, validation criteria, and deployment rules to the shared selection objective.

\subsection{Selection Protocol Adaptation}
\label{app:l2-language-protocol}

We use GPT-5.5 with a default temperature of 0.1 for prompt revision and heuristic construction. The prompt revision loop receives the fixed taxonomy, the current prompt, language-specific evidence, pilot distributions, and named pilot checks. It may revise only the annotation prompt, while the taxonomy, output schema, selection objective, data splits, and pilot thresholds remain fixed. After the prompt is accepted, heuristic construction uses the shared heuristic specification and language evidence. Each resulting heuristic is checked for validity and evaluated on a sampled validation set with a nondegenerate positive rate. The Python file role classification prompt is shown below.

\begin{promptbox}{Python File role classification prompt}
\begin{Verbatim}[breaklines=true,breakanywhere=true,fontsize=\small]
### Task
Classify this Python code into exactly ONE category based on its PRIMARY purpose.

### Category Definitions (Principles First)

**ALGO** - Pure Algorithmic Logic
- Core principle: Code that solves abstract computational problems without real-world I/O
- Key signals: Focus on time/space complexity, uses only stdlib data structures
- Typical: competitive programming, mathematical computations, algorithm implementations
- Boundary: If it reads files, connects to DB, or uses frameworks → NOT ALGO

**WEB** - Network Service Handlers
- Core principle: Code that handles network requests/responses or defines service endpoints
- Key signals: Route decorators, request/response objects, HTTP methods, server setup
- Typical: REST APIs, web views, middleware, WebSocket handlers, RPC services
- Boundary: CLI tools that make HTTP requests are TOOL, not WEB

**TOOL** - Executable Utilities & Libraries
- Core principle: Code meant to be run as scripts OR reused as library components
- Key signals: if __name__ == "__main__", argparse/click, or defines reusable classes/functions
- Typical: CLI tools, automation scripts, crawlers, utility libraries, interface definitions
- Boundary: This is the DEFAULT category when code doesn't fit others

**DATA** - Data & ML Pipelines
- Core principle: Code that transforms, analyzes, or learns from data
- Key signals: ML frameworks, heavy numerical computation, training loops, data transformations
- Typical: model training, feature engineering, ETL, scientific simulations, data analysis
- Scientific libs (spiceypy, astropy, scipy) + numpy/matplotlib → DATA
- Computer vision + ML (face_recognition, cv2 with ML models) → DATA
- Boundary: Simple file parsing without transformation is TOOL, not DATA

**TEST** - Verification Code
- Core principle: Code whose purpose is to verify other code works correctly
- Key signals: test frameworks, assertion statements, test fixtures, mock objects
- Typical: unit tests, integration tests, test utilities
- Boundary: Example/demo code is TOOL, not TEST

**CONFIG** - Pure Declarations
- Core principle: Code that only declares data shapes, settings, or re-exports without logic
- Key signals: No function bodies with logic, only assignments, class definitions without methods
- Typical: settings files, schema definitions, __init__.py with imports, constants
- Boundary: If there's any business logic or complex methods → NOT CONFIG

**EXCLUDE** - Non-classifiable
- Core principle: Code not written by humans OR has no meaningful content
- Key signals: "Generated by", "DO NOT EDIT", migration patterns, empty/stub files
- Auto-generated patterns: _plotly_utils.basevalidators, plotly validators, protobuf *_pb2.py
- Files with ONLY docstrings/comments and no actual code → EXCLUDE
- Typical: auto-generated validators, DB migrations, lock files, empty modules
- IMPORTANT: If you see "_plotly_utils" or "basevalidators" → almost always EXCLUDE

### Decision Process
1. Check EXCLUDE first: Is it auto-generated or empty?
2. Check TEST: Is its purpose to verify other code?
3. Check CONFIG: Is it pure declarations without logic?
4. Check DATA: Does it process/transform/learn from data?
5. Check ALGO: Is it pure algorithmic logic without I/O?
6. Check WEB: Does it handle network requests?
7. Default to TOOL: Utilities, scripts, libraries, anything else

### Generalization Principle
For unfamiliar frameworks or patterns:
- Look at WHAT the code DOES, not which library it uses
- Ask: "What is this code's primary job?" and match to category definitions above

### Output
Reply JSON only: {{"category":"XXX","confidence":0.9}}

### Code
```python
{code}
```
\end{Verbatim}
\end{promptbox}

Candidate examples are sampled in proportion to the rows in each data file using \texttt{seed=42}, producing an annotation sample of 50,000 files for every programming language. The annotator is \texttt{gemini-2.5-flash-nothinking} with temperature 0.1. Files shorter than 50 characters or carrying generated code markers are directly assigned to EXCLUDE. The annotator labels the remaining files, and invalid outputs are removed before training.

By default, selection protocol adaptation evaluates up to five prompt versions on a pilot set of 100 files. Each pilot is checked for valid role names, prediction confidence, and coverage of the taxonomy. A pilot passes when all labels belong to the seven-role taxonomy, the largest role share is at most 0.55, at least four roles appear, the EXCLUDE share is at most 0.15, mean confidence reaches 0.60, and at most 0.30 of predictions have confidence below 0.5. The controller uses the check results to propose the next prompt version. A passing pilot is preferred when the controller proposes no further revision. If five rounds finish without such a pilot, the procedure selects the strongest completed pilot with taxonomy-valid outputs. Completed pilots are ordered by the number of passed checks, mean confidence, category coverage, and distribution balance. When a controller response cannot be parsed or violates the output schema, prompt revision stops and the same selection rule is applied to the completed pilots. At least one completed pilot with taxonomy-valid outputs is required to freeze a protocol. Pilot outcomes and the final protocol choice are recorded for each language.

\subsection{Dual-Cue Supervision Construction}
\label{app:l2-dual-cue-construction}

Dual-cue supervision combines the role label produced by the annotation protocol with a role-conditioned language-specific heuristic cue:

\begin{equation*}
\tilde y_{A,\ell}(x)=
\mathbf{1}[\tilde c(x)=\mathrm{ALGO}\ \lor\ h_\ell(x,\tilde c(x))=1],
\end{equation*}

where $\tilde c(x)$ is the role label assigned by the annotation protocol and $h_\ell(x,\tilde c(x))\in\{0,1\}$ is the decision returned by the language-specific, role-conditioned heuristic protocol. The classifier $C_\ell$ distills the same taxonomy for corpus-scale role prediction and the final EXCLUDE veto. The annotation labels remain the role supervision used to construct the sampled targets for $A_\ell$. Each selection protocol instantiates $h_\ell$ using language-specific function, control-flow, library, and role evidence under a shared heuristic specification. The union of the two cues directly defines the pseudo-labels, which remain unchanged during class balancing.

To obtain stable, disjoint splits independent of input order, we hash each UID with CRC32 and assign the resulting hash buckets to training, validation, and calibration in proportions of 80\%, 10\%, and 10\%. In the Python training set, ALGO positives are oversampled by $2.0\times$, non-ALGO positives by $1.5\times$, and the negative-to-positive ratio is capped at 1.5. When enough negatives are available, positives account for approximately 0.40 of the balanced training rows. Oversampling is applied only to training. Validation and calibration preserve the original labels and class distribution. Validation selects the training epoch, while calibration selects the relevance threshold after the model is frozen.

\subsection{Selection Models}
\label{app:l2-selection-models}

The three selector models use task-specific configuration policies shared across programming languages. The file role classifier and code quality scorer evaluate predefined configuration grids and select a model on their validation splits. The algorithmic relevance classifier uses one configurable default architecture and optimization setting per language, then selects its training epoch on the validation split. The calibration split is reserved for selecting the relevance threshold after the relevance model is frozen. The controller coordinates these validation procedures using the configurations declared in each language protocol.

Qwen3-Embedding-0.6B encodes at most 8,192 tokens per file and produces a shared embedding with $d_{\mathrm{emb}}=1024$ dimensions. Code embeddings are precomputed before model training and treated as fixed inputs during subsequent optimization. The file role classifier and code quality scorer consume this embedding directly. The algorithmic relevance classifier augments it with a 256-dimensional hashed path feature. This design separates corpus encoding from model training, scoring, and selection while allowing relevance scoring to incorporate lightweight repository structure.

\paragraph{File Role Classifier}
The classifier $C_\ell$ is a two-hidden-layer MLP whose hidden blocks each comprise Linear, LayerNorm, GELU, and Dropout. Its default validation grid contains widths $(512,128)$, $(768,256)$, and $(1024,256)$. It includes two $(512,128)$ settings with base dropout and learning rate pairs $(0.30,5\times10^{-4})$ and $(0.25,3\times10^{-4})$, two $(768,256)$ settings with $(0.35,3\times10^{-4})$ and $(0.30,5\times10^{-4})$, and one $(1024,256)$ setting with $(0.35,3\times10^{-4})$. The second hidden block uses 70\% of the listed base dropout rate. All settings predict the seven file roles and use cross-entropy weighted by inverse class frequency, AdamW with weight decay $10^{-4}$, batch size 256, and \texttt{seed=42}. Data are split into 80\% training, 10\% validation, and 10\% held-out evaluation sets. Every configuration is trained for up to 60 epochs and stops after 12 consecutive epochs without an improvement in validation accuracy. Within each setting, the checkpoint with the best validation accuracy is retained. Across settings, validation macro F1 and accuracy select the final configuration in lexicographic order. The selected model is then evaluated once on the held-out split and must reach macro F1 of at least 0.86 and accuracy of at least 0.89 before corpus scoring.

\paragraph{Algorithmic Relevance Classifier}
The algorithmic relevance classifier $A_\ell$ is a two-hidden-layer MLP that maps the combined code and path representation through Linear--ReLU blocks with widths 128 and 64 to a scalar relevance logit. For reproducibility, we detail the path featurization here. Given a repository-relative path, the implementation lowercases it, removes recognized benchmark names, and splits the remaining string on directory and filename separators. For every resulting token longer than three characters, its three-character prefix is added as an additional token. Each token contributes a signed count to one of 256 buckets selected by CRC32 hashing, and collisions are summed. The resulting vector is $\ell_2$ normalized, multiplied by 0.4, and concatenated with the 1024-dimensional code embedding, giving a 1280-dimensional input to $A_\ell$. This representation adds repository-role cues while preventing recognized benchmark names from becoming direct shortcut features. A sigmoid converts the relevance logit into $s_{A,\ell}(x)$. Unlike the file role classifier, $A_\ell$ uses no additional normalization or dropout because it predicts a binary relevance target. Its default optimization setting uses per-example weighted binary cross-entropy against $\tilde y_{A,\ell}(x)$, multiplying weights for ALGO positives by 3.0 and negative weights by 1.5. Adam uses a learning rate of $5\times10^{-3}$, weight decay of $10^{-6}$, batch size 512, at most 20 epochs, and early stopping patience 3. The validation split preserves the original pseudo-label distribution and selects the training epoch by $F_{0.5}$ subject to the validation precision criterion. After the model is frozen, the calibration split supplies the predictions used to select $\tau_{A,\ell}$.

\paragraph{Code Quality Scorer}
Following SwallowCode-v2 and Seed-Coder, we apply the Seed-Coder quality prompt with \texttt{gpt-oss-20b} at temperature 0.1 to obtain integer code quality labels from 0 to 10. Inputs are limited to 8,000 characters, and invalid outputs are removed. The language-specific code quality scorer $Q_\ell$ takes the standardized precomputed code embedding as input. Its shared backbone has two Linear--GELU--LayerNorm--Dropout blocks with hidden widths $d_Q$ and $\max(64,d_Q/2)$. A binary head predicts the zero-score probability $p_{0,\ell}(x)$, while nine ordinal logits $z_{j,\ell}(x)$ model whether a positive score exceeds level $j\in\{1,\ldots,9\}$. We compute the continuous code quality score as

\begin{equation}
s_{Q,\ell}(x)=\bigl(1-p_{0,\ell}(x)\bigr)
\left(1+\sum_{j=1}^{9}\sigma\bigl(z_{j,\ell}(x)\bigr)\right).
\end{equation}

The training objective combines binary cross-entropy for the zero-score head, ordinal binary cross-entropy for positive scores, and Smooth L1 loss on $s_{Q,\ell}(x)$ with weights 1, 1, and 0.2. The default configuration grid consists of predefined settings drawn from $d_Q\in\{384,512,768,1024\}$, dropout rates in $\{0.10,0.15,0.20\}$, and learning rates in $\{5\times10^{-4},10^{-3}\}$. Models use AdamW with weight decay $10^{-5}$, batch size 256, at most 80 epochs, patience 10, an 80/10/10 split, and \texttt{seed=42}. Model configurations are ranked by validation quadratic weighted kappa, mean absolute error, and the ratios of errors within two and above five points, with rank weights 0.35, 0.30, 0.15, and 0.20. The selected model is then evaluated on the held-out split. For full corpus selection, the quality threshold is fixed at $\tau_Q=3$ to remove the low-score region while retaining medium-to-high-scoring files.

\begin{figure}[t]
    \centering
    \includegraphics[width=0.5\textwidth]{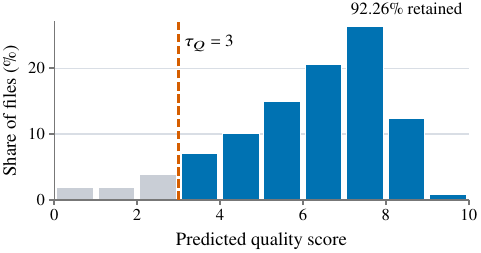}
    \caption{Predicted code quality score distribution in the Python candidate pool. The shared threshold $\tau_Q=3$ removes the low-score tail and retains 92.26\% of files.}
    \label{fig:quality_score_distribution}
\end{figure}

\textbf{The quality threshold acts as an auxiliary veto.} As shown in Figure~\ref{fig:quality_score_distribution}, 92.26\% of Python candidate files satisfy $s_{Q,\mathrm{py}}(x)\ge3$. The threshold therefore removes a limited low-score tail, while algorithmic relevance remains the main criterion for corpus selection.

\subsection{Corpus Scoring and Full-Corpus Selection}
\label{app:l2-threshold-selection}

After the three selection models are frozen, they score the full candidate corpus. The resulting scores are used to evaluate the language-specific relevance threshold $\tau_{A,\ell}$. The quality threshold remains fixed at $\tau_Q=3$, and EXCLUDE predictions are removed according to the selection function defined in Section~\ref{sec:l2-selector}. For programming language $\ell$, we construct the following feasible set over the grid $0.50\leq t\leq0.85$ with a step size of 0.01:

\begin{equation}
\mathcal T_\ell=
\left\{
t:\ \operatorname{Prec}_{\mathrm{overall}}(t)\geq0.85,
\ \operatorname{Rec}_{\mathrm{ALGO}}(t)\geq0.80
\right\}.
\end{equation}

For a grid value $t$, the retained calibration files are those with $s_{A,\ell}(x)\geq t$. $\operatorname{Prec}_{\mathrm{overall}}(t)$ is the precision of these files against their pseudo-labels. $\operatorname{Rec}_{\mathrm{ALGO}}(t)$ is the fraction of ALGO files in the calibration split that meet the threshold. A threshold belongs to $\mathcal T_\ell$ when both conditions hold. Feasible thresholds are ordered by $F_{0.5}$, overall precision, ALGO recall, and threshold value, all in descending order. If no threshold is feasible, selection stops until the protocol or model configuration is revised. Otherwise, the selected threshold and the frozen models define the corpus-level selection rule. This procedure selects $\tau_{A,\mathrm{py}}=0.85$ for Python.

Full-corpus selection applies the frozen threshold and auxiliary conditions to the model scores. The resulting corpus is the set of files that satisfy the relevance, role, and quality conditions.

\subsection{L2 Data Statistics}
\label{app:l2-data-statistics}

The 11 language-specific subsets of UltraData-Code-L2 correspond to the 11 language suffix partitions inherited from UltraData-Code-L1. The reported corpus composition in Table~\ref{tab:l2-language-statistics} is measured on the experimental snapshot before the decontamination step. The language-level retention rates are effective rates after all L2 selection operations and the language-specific downsampling. JavaScript and PHP additionally apply language-specific downsampling due to their substantially larger token volumes, which is therefore already reflected in the reported retention rates and the approximately 400B-token total. The effective size of the public release is determined by the files and tokens actually made available.

\begin{table}[htbp]
    \centering
    \caption{Language-level effective retention and token composition of the UltraData-Code-L2 experimental snapshot.}
    \label{tab:l2-language-statistics}
    \vspace{1mm}
    \begin{tabular}{@{}lrrr@{}}
        \toprule
        \textbf{Language} & \textbf{Retention} & \textbf{Tokens} & \textbf{Share} \\
        \midrule
        Python & 12.96\% & 102.57B & 25.23\% \\
        Java & 13.80\% & 57.99B & 14.26\% \\
        C++ & 28.82\% & 31.94B & 7.86\% \\
        C\# & 14.27\% & 49.70B & 12.22\% \\
        Go & 13.30\% & 5.32B & 1.31\% \\
        Ruby & 18.18\% & 6.43B & 1.58\% \\
        JavaScript & 9.18\% & 86.69B & 21.32\% \\
        PHP & 5.42\% & 54.56B & 13.42\% \\
        R & 20.88\% & 1.33B & 0.33\% \\
        Rust & 25.44\% & 5.33B & 1.31\% \\
        Shell & 18.84\% & 4.72B & 1.16\% \\
        \midrule
        \textbf{Total} & \textbf{12.23\%} & \textbf{406.58B} & \textbf{100\%} \\
        \bottomrule
    \end{tabular}
\end{table}

\subsection{Selection Analysis}
\label{app:l2-selection-analysis}

\begin{figure}[htbp]
    \centering
    \begin{subfigure}[t]{0.46\textwidth}
        \centering
        \includegraphics[width=0.98\linewidth]{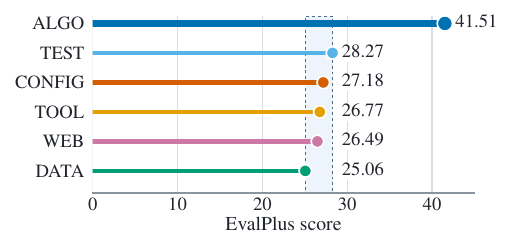}
        \caption{Role probe results on EvalPlus after continual pre-training with 10B tokens.}
        \label{fig:analysis-file-role-probe}
    \end{subfigure}\hfill
    \begin{subfigure}[t]{0.46\textwidth}
        \centering
        \includegraphics[width=0.98\linewidth]{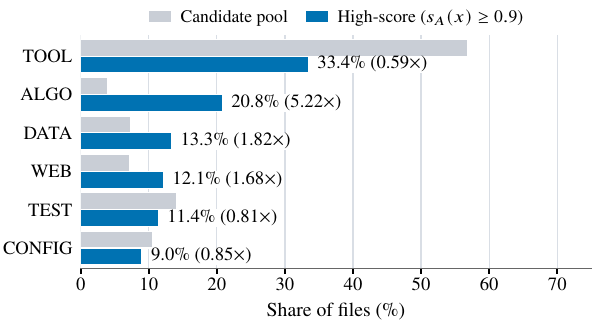}
        \caption{Python file role composition before and after relevance scoring.}
        \label{fig:analysis-role-composition}
    \end{subfigure}
    \caption{Evidence for algorithmic relevance selection across file roles. The shaded band in (a) covers roles other than ALGO. Parentheses in (b) show enrichment relative to the candidate pool.}
    \label{fig:analysis-role-evidence}
\end{figure}

\paragraph{ALGO provides a concentrated anchor for relevance learning.}
\label{app:analysis-algo-anchor}
We partition UltraData-Code-L1-py into six eligible role corpora using the frozen predictions of $C_{\mathrm{py}}$ and omit files predicted as EXCLUDE. Each role corpus supports a separate 10B-token continual pre-training run with the same settings and EvalPlus evaluation as the main experiments. The predicted file role is therefore the only experimental factor that differs across runs.
As shown in Figure~\ref{fig:analysis-file-role-probe}, ALGO accounts for 3.70\% of UltraData-Code-L1-py files and 1.68\% of its estimated tokens, yet training on this role produces the strongest EvalPlus result in the controlled role probe. This combination of scarcity and downstream utility makes ALGO a precise source of supervision. It identifies files centered on algorithm implementation while leaving room for relevant computation that appears inside files with other primary roles. 
Figure~\ref{fig:analysis-role-composition} shows the role composition after algorithmic relevance scoring. Among files outside EXCLUDE, the subset with $s_{A}(x)\ge0.9$ most strongly enriches ALGO and also enriches DATA and WEB. TOOL becomes less dominant, while TEST and CONFIG remain present without enrichment. The score therefore concentrates algorithmic content while retaining examples from multiple file roles.

\paragraph{Dual-cue supervision extends the ALGO anchor across file roles.}

\begin{figure}[htbp]
    \centering
    \begin{subfigure}[t]{0.46\textwidth}
        \centering
        \includegraphics[width=0.6\linewidth]{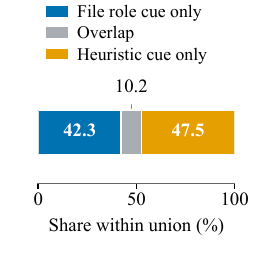}
        \caption{Cue source composition.}
        \label{fig:l2-dual-cue-source}
    \end{subfigure}\hfill
    \begin{subfigure}[t]{0.46\textwidth}
        \centering
        \includegraphics[width=0.6\linewidth]{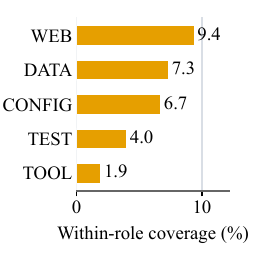}
        \caption{Coverage outside ALGO.}
        \label{fig:l2-dual-cue-coverage}
    \end{subfigure}
    \caption{Coverage of the two supervision cues in the Python relevance sample, excluding EXCLUDE.}
    \label{fig:analysis-dual-cue-evidence}
\end{figure}

The heuristic cue identifies algorithmic evidence in DATA, WEB, TOOL, TEST, and CONFIG files, complementing the files whose primary role is ALGO. As shown in Figure~\ref{fig:analysis-dual-cue-evidence}, the two cues contribute complementary positive samples. Their union yields 90.5\% more positive Python examples than the ALGO cue alone, and the added supervision spans every examined role other than ALGO. The relevance classifier can therefore learn from varied expressions of problem-solving logic instead of treating file role as the final selection rule.

\subsection{UltraData-Stack-v3-L2 Construction and Statistics}
\label{app:l2-ultra-stack-v3}

UltraData-Stack-v3-L2 is constructed from the independently sourced Stack-v3-train corpus. Based on quality-inspection results, the cleaning stage removes records whose full text consists of URLs, records below the minimum length threshold, and records containing garbled or malformed text. After this cleaning step, we run the same language-adaptive L2 selector used for UltraData-Code-L2, including file role and algorithmic relevance labeling, the EXCLUDE veto, and the shared quality constraint. The selector covers the same 11 programming languages, while the relevance threshold is calibrated separately for each language on the cleaned Stack-v3-train source. The resulting filtering statistics are reported in Table~\ref{tab:ultra-stack-v3-statistics}.

\begin{table}[htbp]
    \centering
    \caption{Language-level construction statistics for UltraData-Stack-v3-L2 after cleaning and L2 selection.}
    \label{tab:ultra-stack-v3-statistics}
    \vspace{1mm}
    \small
    \begin{tabular}{lrrrrr}
        \toprule
        \textbf{Language} & \textbf{After cleaning} & \textbf{After selection} & \textbf{Retention} & \textbf{Token Share} & \textbf{Tokens} \\
        \midrule
        Python & 96{,}961{,}008 & 24{,}671{,}724 & 25.44\% & 15.86\% & 38.94B \\
        Java & 165{,}346{,}828 & 41{,}963{,}467 & 25.38\% & 28.76\% & 70.60B \\
        JavaScript & 110{,}967{,}790 & 14{,}521{,}755 & 13.09\% & 12.92\% & 31.73B \\
        C++ & 67{,}981{,}897 & 17{,}990{,}102 & 26.46\% & 12.78\% & 31.38B \\
        PHP & 65{,}565{,}149 & 5{,}500{,}141 & 8.39\% & 10.26\% & 25.19B \\
        Ruby & 32{,}667{,}620 & 6{,}389{,}441 & 19.56\% & 2.39\% & 5.86B \\
        Shell & 26{,}704{,}436 & 5{,}045{,}813 & 18.90\% & 8.83\% & 21.67B \\
        C\# & 17{,}309{,}141 & 3{,}548{,}274 & 20.50\% & 3.64\% & 8.93B \\
        Go & 29{,}683{,}899 & 3{,}138{,}105 & 10.57\% & 1.65\% & 4.04B \\
        Rust & 11{,}674{,}463 & 2{,}907{,}231 & 24.90\% & 2.13\% & 5.24B \\
        R & 7{,}640{,}182 & 1{,}456{,}974 & 19.07\% & 0.78\% & 1.92B \\
        \midrule
        \textbf{Total} & \textbf{632{,}502{,}413} & \textbf{127{,}133{,}027} & \textbf{20.10\%} & \textbf{100\%} & \textbf{245.50B} \\
        \bottomrule
    \end{tabular}
\end{table}

%% file: sections/apx-C-L3-Synthesis-Details.tex
\section{L3 Synthesis Details}
\label{app:l3-synthesis-details}

In this appendix, we document the L3 seed scope, generation prompts, multilingual data scale, serialization templates, and training set definitions used in the L3 experiments.

\subsection{Synthesis Configuration}

The seeds for UltraData-Code-L3 are drawn from files in UltraData-Code-L2 that are assigned to ALGO. L3 code generation uses DeepSeek-v4-flash in no-think mode. Given one source snippet, a shared prompt schema with language-specific syntax and interface instructions requests a single structured response with four sections: \emph{task}, \emph{analysis}, \emph{solution}, and \emph{test}. These sections define a standalone programming exercise, explain its algorithm and complexity, request a self-contained reference implementation, and supply test-oriented supervision. The prompt preserves the seed's principal computational intent and input-output behavior while removing repository-specific context.

The general synthesis prompt template, as shown below, uses \texttt{\{Language\}}, \texttt{\{LanguageTag\}}, and \texttt{\{Code\}} as placeholders for the target programming language, its Markdown code fence identifier, and the seed content.

\begin{promptbox}{General Synthesis Prompt}
\begin{Verbatim}[breaklines=true,breakanywhere=true,fontsize=\small]
You are a teaching assistant creating a standalone {Language} programming task inspired by a 
given code snippet. The resulting task must be self-contained and suitable as an independent 
programming exercise. Provide a high-quality response that includes analysis, a reference 
solution, and test code.

[Code Snippet]
```{LanguageTag}
{Code}
```

Your response must contain exactly these sections:
[Task]
{Create an independent and detailed {Language} programming task in one paragraph.}
[Analysis]
{Explain the solution approach clearly, including the main algorithm, important edge cases, and 
time and space complexity.}
[Solution]
{Write a high-quality, self-contained {Language} implementation, including idiomatic 
documentation with necessary information. Output code only. Do not include any entry-point 
wrapper (e.g., `main`, `public static void main`, or an equivalent entry point).}
[Test]
{Provide 1-10 runnable checks to verify the solution. Each test case must be a standalone 
assertion statement or an equivalent assertion in {Language}.}

Example output format (illustrative only, do not copy content):
[Task]
Given a list of integers, write a function that returns a new list where each element is the 
cumulative sum up to that position.
[Analysis]
We need to iterate through the input list while maintaining a running total. For each number, 
add it to the running total and append the total to the result list. This guarantees each 
output element equals the sum of all previous and current elements.
[Solution]
```{LanguageTag}
{Provide an idiomatic {Language} implementation here.}
```
[Test]
```{LanguageTag}
{Provide standalone assertions or equivalent runnable checks in {Language} here.}
```
\end{Verbatim}
\end{promptbox}

The language-specific prompts instantiate the three placeholders in the shared template and specialize only interface and execution requirements. The Python specialization, as shown below, sets the language and code-fence placeholders, requests a self-contained function with a docstring and no entry-point wrapper, and asks for 1--10 standalone assertion checks. The other languages follow the same adaptation pattern.

\begin{promptbox}{Python Synthesis Prompt}
\begin{Verbatim}[breaklines=true,breakanywhere=true,fontsize=\small]
You are a teaching assistant creating a standalone Python programming task inspired by a given 
code snippet. The resulting task must be self-contained and suitable as an independent 
programming exercise. Provide a high-quality response that includes analysis, a reference 
solution, and test code.

[Code Snippet]
```python
{Code}
```

Your response must contain exactly these sections:
[Task]
{Create an independent and detailed Python programming task in one paragraph.}
[Analysis]
{Explain the solution approach clearly, including the main algorithm, important edge cases, and 
time and space complexity.}
[Solution]
{Write a high-quality, self-contained Python function implementation, including a docstring with 
necessary information. Output code only. Do not include any entry-point wrapper (e.g., `if 
__name__ == "__main__":`, `main()` guards)}
[Test]
{Provide 1-10 runnable checks to verify the solution. Each test case must be a standalone 
assertion statement.}

Example output format (illustrative only, do not copy content):
[Task]
Given a list of integers, write a program to return a new list where each element is the 
cumulative sum up to that position.
[Analysis]
We need to iterate through the input list while maintaining a running total. For each number, 
add it to the running total and append the total to the result list. This guarantees each 
output element equals the sum of all previous and current elements.
[Solution]
```python
def cumulative_sum(nums):
    """Return the cumulative sum at each position in `nums`.

    Args:
        nums: A list of integers.

    Returns:
        A new list in which each element is the sum of all values
        from the beginning of `nums` through the corresponding position.
    """
    total = 0
    result = []
    for number in nums:
        total += number
        result.append(total)
    return result
```
[Test]
```python
assert cumulative_sum([]) == []
assert cumulative_sum([5]) == [5]
assert cumulative_sum([1, 2, 3, 4]) == [1, 3, 6, 10]
```
\end{Verbatim}
\end{promptbox}

\subsection{L3 Data Statistics}

Table~\ref{tab:l3-language-token-scale} summarizes the scale of L3 construction before experimental serialization for each of the 11 programming languages. The ALGO seeds column denotes the eligible ALGO files from L2, while the Synthesis column counts all generation attempts, including outputs that later fail parsing. The Samples column counts successfully parsed records containing the four required fields. Content measures the combined token volume of task and solution, whereas Full Content includes task, analysis, solution, and test.

Overall, 81.30 million ALGO seeds yield 81.25 million generation attempts and 81.21 million successfully parsed records. Among the attempted generations, 99.95\% satisfy the required four-field structure. The task and solution fields contain 63.06B tokens, while all four fields contain 156.37B tokens in total. The latter provides the basis for the approximately 150B-token scale reported in the main paper. The main experiments use a fixed 10B-token training budget. Within this budget, L3 records are serialized using task and solution only unless otherwise stated.

\begin{table}[htbp]
    \centering
    \caption{UltraData-Code-L3 construction scale and token volume by programming language. The Synthesis column includes generation attempts that fail parsing. The Samples column counts records containing all four required fields. Content counts task and solution tokens. Full Content counts tokens in all four fields.}
    \label{tab:l3-language-token-scale}
    \vspace{1mm}
    \setlength{\tabcolsep}{3pt}
    \begin{tabular}{lrrrrr}
        \toprule
        \textbf{Language} & \textbf{ALGO seeds} & \textbf{Synthesis} & \textbf{Samples} & \textbf{Content} & \textbf{Full Content} \\
        \midrule
        Python & 20,813,025 & 20,758,682 & 20,750,202 & 16.24B & 38.88B \\
        Java & 21,419,871 & 21,419,746 & 21,415,405 & 16.39B & 38.38B \\
        C++ & 22,611,690 & 22,611,024 & 22,600,686 & 18.00B & 49.09B \\
        C\# & 3,615,964 & 3,615,945 & 3,614,454 & 3.11B & 7.43B \\
        Go & 1,962,358 & 1,962,336 & 1,948,148 & 1.41B & 3.48B \\
        Ruby & 847,079 & 847,063 & 846,872 & 0.42B & 1.21B \\
        JavaScript & 6,132,686 & 6,132,605 & 6,131,527 & 2.93B & 7.97B \\
        PHP & 2,307,612 & 2,307,581 & 2,306,009 & 3.23B & 6.64B \\
        R & 304,102 & 304,088 & 304,016 & 0.30B & 0.66B \\
        Rust & 923,270 & 923,254 & 922,564 & 0.81B & 1.99B \\
        Shell & 365,362 & 365,360 & 365,222 & 0.22B & 0.64B \\
        \midrule
        \textbf{Total} & \textbf{81,303,019} & \textbf{81,247,684} & \textbf{81,205,105} & \textbf{63.06B} & \textbf{156.37B} \\
        \bottomrule
    \end{tabular}
\end{table}

\subsection{Serialization Configurations and Templates}
\label{app:l3-serialization-templates}

We treat content generation and training text representation as separate stages. Each released record preserves four generated fields comprising task, analysis, solution, and test. In the main UltraData-Code-L3 setting, each training sample contains only task and solution. To maintain presentation diversity, examples are randomly assigned to \emph{code comment}, \emph{no fence}, and \emph{Markdown} formats with proportions of 50\%, 25\%, and 25\%, respectively. The two fields are separated by one or two line breaks.

The \emph{code comment} format represents the task as a comment or string literal appropriate to the target language before the implementation. The Python example below uses a triple-quoted string. The \emph{no fence} format directly concatenates the task and solution without a Markdown fence. The \emph{Markdown} format places the task before the solution and encloses the implementation in a code fence tagged with the target language. The following examples illustrate these three serialization formats. Ellipses denote task-specific text or function implementations.

\begin{promptbox}{task + solution in code comment format}
\begin{verbatim}
"""
Using the provided BinarySearchTree class as a foundation, write a standalone
Python function find_kth_smallest(root, k) that takes the root node of a binary
search tree and an integer k (1-indexed), and returns the k-th smallest value
in the tree. If k is less than 1 or greater than the total number of nodes in
the tree, raise a ValueError with an appropriate message. The function must
not modify the tree structure and should work efficiently even for large trees.
"""
def find_kth_smallest(root, k):
    """Return the k-th smallest value in a binary search tree.
    ...
    """
    ...
    return result
\end{verbatim}
\end{promptbox}

\begin{promptbox}{task + solution with no fence}
\begin{verbatim}
Write a Python function mayan_value(tokens) that takes a list of strings
representing Mayan-style tokens ...
def mayan_value(tokens):
    """Compute the base-20 value of a list of Mayan-like tokens.
    ...
    """
    ...
    return total
\end{verbatim}
\end{promptbox}

\begin{promptbox}{task + solution in Markdown format}
\begin{verbatim}
Create a Python function called `gladiator_expenses` that takes the number of
lost fights ...
```python
def gladiator_expenses(lost_fights, helmet_price, sword_price,
                       shield_price, armor_price):
    """Calculate total gladiator repair expenses.
    ...
    """
    ...
    return f"Gladiator expenses: {total_cost:.2f} aureus"
```
\end{verbatim}
\end{promptbox}

The four-field variant additionally serializes analysis and test. Its examples are distributed equally across four templates using \emph{standard order}, \emph{commented standard order}, \emph{solution first order}, and \emph{test first order}. These templates differ in field order and comment placement, while preserving the same generated content.

\begin{promptbox}{standard order}
\begin{verbatim}
Write a Python function `break_vigenere(...)` ... Assume all inputs are valid
and the key length is exactly 12.

The approach is straightforward:
1. For each offset `i` from 0 to 11, ...
...
Time complexity: O(n) ... Space complexity: O(n) ...

from collections import Counter
from itertools import cycle

def break_vigenere(ciphertext_numbers, ciphertext_ascii, key_length=12):
    """Decrypt a Vigenere-style cipher ..."""
    ...
    return ''.join(decrypted_chars)

# Test 1: Simple known key and short plaintext
key = [10, 20, 30, ...]
assert break_vigenere(cipher_nums, cipher_ascii) == plain
\end{verbatim}
\end{promptbox}

\begin{promptbox}{commented standard order}
\begin{verbatim}
Write a Python function `merge_sorted_with_infinity` that takes two sorted
lists ...

# The core challenge is merging two pre-sorted sequences while respecting ...
# Time complexity is O(n+m) ...

def merge_sorted_with_infinity(list1, list2):
    """Merge two sorted lists (which may contain Infinity and -Infinity) ..."""
    ...
    return result

class InfinityType:
    ...
assert merge_sorted_with_infinity([], []) == []
assert merge_sorted_with_infinity([1, 3, 5], [2, 4, 6]) == [1, 2, 3, 4, 5, 6]
\end{verbatim}
\end{promptbox}

\begin{promptbox}{solution-first order}
\begin{verbatim}
"""
Using the provided BinarySearchTree class as a foundation, write a standalone
Python function `find_kth_smallest(root, k)` ...
"""
def find_kth_smallest(root, k):
    """Return the k-th smallest value in a binary search tree."""
    ...
    return result

# Helper to build a BST from a list of values
def build_bst(values):
    ...

assert find_kth_smallest(root, 1) == 2
...

# The most straightforward approach is to perform an in-order traversal ...
# Space complexity is O(h) for recursion stack ...
\end{verbatim}
\end{promptbox}

\begin{promptbox}{test-first order}
\begin{verbatim}
Write a Python function `getKthFromEnd(head, k)` that takes the head of a
singly linked list ...

# Helper to build list from list of values
def build_list(values):
    ...

assert getKthFromEnd(None, 1) is None
assert getKthFromEnd(build_list([1, 2, 3, 4, 5]), 2).val == 4
...

class ListNode:
    def __init__(self, data=0, next_node=None):
        ...

def getKthFromEnd(head: ListNode, k: int) -> ListNode:
    """Return the k-th node from the end of a singly linked list."""
    ...
    return behind

# The key is to use two pointers: `ahead` and `behind` ...
# Space complexity: O(1) extra space.
\end{verbatim}
\end{promptbox}

%% file: sections/apx-D-Multilingual-Evaluation-Details.tex
\clearpage
\section{Detailed Multilingual Evaluation Results}
\label{app:multipl-e-results}

The main multilingual evaluation reports macro averages in Table~\ref{tab:l2-multilingual-main-results}. This appendix provides the corresponding pass@1 scores for each of the ten target languages in MultiPL-E, separately for HumanEval and MBPP in Tables~\ref{tab:multipl-e-humaneval-details} and~\ref{tab:multipl-e-mbpp-details}. \textit{AVG.} denotes the unweighted mean computed from the original scores before rounding. We abbreviate UltraData-Code as UDC, UltraData-Stack-v3-L2 as UDS-v3-L2, and CodeAlchemy as CA, while A\&T denotes the variant containing analysis and test fields.

\begin{table}[H]
    \centering
    \captionsetup{font=footnotesize,skip=2pt}
    \caption{Pass@1 scores for each language on MultiPL-E HumanEval after 10B tokens of continual pre-training. Each L3 dataset is mixed with UDC-L2 in equal token proportions.}
    \label{tab:multipl-e-humaneval-details}
    \scriptsize
    \renewcommand{\arraystretch}{0.92}
    \setlength{\tabcolsep}{2pt}
    \resizebox{0.94\textwidth}{!}{%
    \begin{tabular}{clccccccccccc}
        \toprule
        \textbf{Level} & \textbf{Dataset} & \textbf{C++} & \textbf{C\#} & \textbf{Go} & \textbf{Java} & \textbf{Ruby} & \textbf{JS} & \textbf{PHP} & \textbf{R} & \textbf{Rust} & \textbf{Shell} & \textbf{AVG.} \\
        \midrule
        \multirow{4}{*}{\textbf{L1}}
        & Stack-v2 & 18.63 & 7.59 & 5.84 & 5.70 & 8.70 & 8.70 & 9.94 & 7.45 & 3.21 & 1.90 & 7.77 \\
        & RefineCode & 18.01 & 13.29 & 7.79 & 13.92 & 13.04 & 20.50 & 12.42 & 6.21 & 12.82 & 5.70 & 12.37 \\
        & Stack-v3-train & 16.15 & 12.66 & 11.04 & 22.15 & 13.04 & 19.25 & 12.42 & 6.21 & 9.62 & 3.16 & 12.57 \\
        & \cellcolor{cellHighlight}UDC-L1 & \cellcolor{cellHighlight}16.15 & \cellcolor{cellHighlight}9.49 & \cellcolor{cellHighlight}7.14 & \cellcolor{cellHighlight}15.19 & \cellcolor{cellHighlight}13.04 & \cellcolor{cellHighlight}14.29 & \cellcolor{cellHighlight}11.80 & \cellcolor{cellHighlight}7.45 & \cellcolor{cellHighlight}9.62 & \cellcolor{cellHighlight}3.80 & \cellcolor{cellHighlight}10.80 \\
        \midrule
        \multirow{3}{*}{\textbf{L2}}
        & Stack-Edu & 18.63 & 12.03 & 3.25 & 8.86 & 16.77 & 19.25 & 14.29 & 8.07 & 11.54 & 5.06 & 11.78 \\
        & \cellcolor{cellHighlight}UDS-v3-L2 & \cellcolor{cellHighlight}21.12 & \cellcolor{cellHighlight}19.62 & \cellcolor{cellHighlight}14.29 & \cellcolor{cellHighlight}23.42 & \cellcolor{cellHighlight}19.25 & \cellcolor{cellHighlight}26.71 & \cellcolor{cellHighlight}20.50 & \cellcolor{cellHighlight}11.80 & \cellcolor{cellHighlight}18.59 & \cellcolor{cellHighlight}6.33 & \cellcolor{cellHighlight}18.16 \\
        & \cellcolor{cellHighlight}UDC-L2 & \cellcolor{cellHighlight}19.88 & \cellcolor{cellHighlight}11.39 & \cellcolor{cellHighlight}10.39 & \cellcolor{cellHighlight}17.72 & \cellcolor{cellHighlight}17.39 & \cellcolor{cellHighlight}24.22 & \cellcolor{cellHighlight}18.63 & \cellcolor{cellHighlight}8.07 & \cellcolor{cellHighlight}10.90 & \cellcolor{cellHighlight}6.96 & \cellcolor{cellHighlight}14.56 \\
        \midrule
        \multirow{5}{*}{\textbf{L3}}
        & CA-QA & 21.74 & 15.82 & 16.23 & 20.25 & 14.91 & 22.98 & 21.12 & 8.07 & 17.31 & 4.43 & 16.29 \\
        & CA-Enhance & 22.98 & 14.56 & 13.64 & 25.32 & 19.25 & 25.47 & 13.66 & 9.32 & 10.90 & 5.70 & 16.08 \\
        & CA-Dialogue & 18.01 & 12.03 & 5.84 & 18.99 & 14.91 & 21.12 & 7.45 & 7.45 & 9.62 & 3.80 & 11.92 \\
        & \cellcolor{cellHighlight}UDC-L3 (A\&T) & \cellcolor{cellHighlight}31.68 & \cellcolor{cellHighlight}17.09 & \cellcolor{cellHighlight}20.78 & \cellcolor{cellHighlight}20.89 & \cellcolor{cellHighlight}24.84 & \cellcolor{cellHighlight}29.19 & \cellcolor{cellHighlight}30.43 & \cellcolor{cellHighlight}12.42 & \cellcolor{cellHighlight}19.23 & \cellcolor{cellHighlight}7.59 & \cellcolor{cellHighlight}21.42 \\
        & \cellcolor{cellHighlight}UDC-L3 & \cellcolor{cellHighlight}31.68 & \cellcolor{cellHighlight}18.35 & \cellcolor{cellHighlight}21.43 & \cellcolor{cellHighlight}31.01 & \cellcolor{cellHighlight}31.68 & \cellcolor{cellHighlight}36.65 & \cellcolor{cellHighlight}26.71 & \cellcolor{cellHighlight}17.39 & \cellcolor{cellHighlight}25.00 & \cellcolor{cellHighlight}9.49 & \cellcolor{cellHighlight}24.94 \\
        \bottomrule
    \end{tabular}}
\end{table}

\begin{table}[H]
    \centering
    \captionsetup{font=footnotesize,skip=2pt}
    \caption{Pass@1 scores for each language on MultiPL-E MBPP after 10B tokens of continual pre-training. Each L3 dataset is mixed with UDC-L2 in equal token proportions.}
    \label{tab:multipl-e-mbpp-details}
    \scriptsize
    \renewcommand{\arraystretch}{0.92}
    \setlength{\tabcolsep}{2pt}
    \resizebox{0.94\textwidth}{!}{%
    \begin{tabular}{clccccccccccc}
        \toprule
        \textbf{Level} & \textbf{Dataset} & \textbf{C++} & \textbf{C\#} & \textbf{Go} & \textbf{Java} & \textbf{Ruby} & \textbf{JS} & \textbf{PHP} & \textbf{R} & \textbf{Rust} & \textbf{Shell} & \textbf{AVG.} \\
        \midrule
        \multirow{4}{*}{\textbf{L1}}
        & Stack-v2 & 26.45 & 13.21 & 10.96 & 22.02 & 16.37 & 16.88 & 19.14 & 11.08 & 4.80 & 1.83 & 14.27 \\
        & RefineCode & 27.20 & 20.73 & 14.97 & 25.39 & 20.91 & 30.23 & 28.46 & 8.82 & 23.16 & 12.83 & 21.27 \\
        & Stack-v3-train & 28.97 & 22.02 & 22.19 & 27.20 & 23.43 & 32.24 & 25.94 & 14.11 & 14.69 & 9.42 & 22.02 \\
        & \cellcolor{cellHighlight}UDC-L1 & \cellcolor{cellHighlight}21.41 & \cellcolor{cellHighlight}18.65 & \cellcolor{cellHighlight}14.44 & \cellcolor{cellHighlight}25.65 & \cellcolor{cellHighlight}19.90 & \cellcolor{cellHighlight}30.98 & \cellcolor{cellHighlight}23.17 & \cellcolor{cellHighlight}12.34 & \cellcolor{cellHighlight}21.75 & \cellcolor{cellHighlight}8.64 & \cellcolor{cellHighlight}19.69 \\
        \midrule
        \multirow{3}{*}{\textbf{L2}}
        & Stack-Edu & 31.74 & 22.80 & 11.50 & 28.76 & 20.40 & 35.77 & 26.95 & 13.60 & 24.29 & 13.09 & 22.89 \\
        & \cellcolor{cellHighlight}UDS-v3-L2 & \cellcolor{cellHighlight}33.75 & \cellcolor{cellHighlight}25.91 & \cellcolor{cellHighlight}21.93 & \cellcolor{cellHighlight}33.16 & \cellcolor{cellHighlight}32.24 & \cellcolor{cellHighlight}35.77 & \cellcolor{cellHighlight}30.23 & \cellcolor{cellHighlight}17.88 & \cellcolor{cellHighlight}30.23 & \cellcolor{cellHighlight}12.30 & \cellcolor{cellHighlight}27.34 \\
        & \cellcolor{cellHighlight}UDC-L2 & \cellcolor{cellHighlight}31.99 & \cellcolor{cellHighlight}24.87 & \cellcolor{cellHighlight}25.13 & \cellcolor{cellHighlight}30.31 & \cellcolor{cellHighlight}27.46 & \cellcolor{cellHighlight}37.28 & \cellcolor{cellHighlight}30.98 & \cellcolor{cellHighlight}20.91 & \cellcolor{cellHighlight}22.32 & \cellcolor{cellHighlight}10.73 & \cellcolor{cellHighlight}26.20 \\
        \midrule
        \multirow{5}{*}{\textbf{L3}}
        & CA-QA & 31.99 & 17.88 & 21.39 & 29.79 & 28.97 & 36.78 & 27.20 & 15.87 & 28.53 & 11.78 & 25.02 \\
        & CA-Enhance & 30.48 & 23.06 & 22.73 & 28.50 & 28.72 & 36.78 & 30.98 & 17.88 & 21.47 & 10.99 & 25.16 \\
        & CA-Dialogue & 31.74 & 20.73 & 10.70 & 25.65 & 26.95 & 34.76 & 22.92 & 12.85 & 22.03 & 10.47 & 21.88 \\
        & \cellcolor{cellHighlight}UDC-L3 (A\&T) & \cellcolor{cellHighlight}33.00 & \cellcolor{cellHighlight}27.20 & \cellcolor{cellHighlight}32.89 & \cellcolor{cellHighlight}34.72 & \cellcolor{cellHighlight}29.47 & \cellcolor{cellHighlight}39.80 & \cellcolor{cellHighlight}36.27 & \cellcolor{cellHighlight}18.39 & \cellcolor{cellHighlight}24.29 & \cellcolor{cellHighlight}15.45 & \cellcolor{cellHighlight}29.15 \\
        & \cellcolor{cellHighlight}UDC-L3 & \cellcolor{cellHighlight}36.52 & \cellcolor{cellHighlight}31.35 & \cellcolor{cellHighlight}36.10 & \cellcolor{cellHighlight}36.79 & \cellcolor{cellHighlight}39.29 & \cellcolor{cellHighlight}41.06 & \cellcolor{cellHighlight}37.03 & \cellcolor{cellHighlight}20.40 & \cellcolor{cellHighlight}29.10 & \cellcolor{cellHighlight}12.04 & \cellcolor{cellHighlight}31.97 \\
        \bottomrule
    \end{tabular}}
\end{table}